\documentclass[acmlarge]{acmart}

\renewcommand\footnotetextcopyrightpermission[1]{} %
\usepackage{expl3}
\expandafter\def\csname ver@l3regex.sty\endcsname{} 
\usepackage{booktabs} %
\usepackage{amsmath}
\usepackage{xspace}
\usepackage{url}
\usepackage{subfigure}
\usepackage{multirow}
\usepackage{array}
\usepackage{algorithm}
\usepackage{algpseudocode}
\usepackage[normalem]{ulem}
\usepackage{gensymb}
\usepackage{caption}
\usepackage{graphicx}
\usepackage{geometry}
\usepackage[T1]{fontenc}
\usepackage{breakurl}
\usepackage{balance}
\usepackage{cryptocode}
\usepackage{tikz}
\usetikzlibrary{matrix,shapes,arrows,positioning,chains, calc}
\usepackage{tabu}
\usepackage{amssymb}
\usepackage{subfiles}
\usepackage{setspace}
\usepackage[flushleft]{threeparttable}
\setcopyright{none}

\newcolumntype{L}{>{\centering\arraybackslash}m{3cm}}
\newcounter{linecounter}

\newcommand{\altname}{\textit{LaNet}\xspace}
\newcommand{\block}[1]{}                %

\def\centerhack#1{\hbox to 0pt{\hss\footnotesize #1\hss}}

\algnewcommand{\LineComment}[1]{\State \(\triangleright\) #1}

\title{\altname: %
Real-time Lane Identification by Learning Road Surface Characteristics from Accelerometer Data}

\author{Madhumitha Harishankar}
\email{mharisha@andrew.cmu.edu}
\affiliation{%
  \institution{Carnegie Mellon University}
}
\author{Jun Han}
\email{junhan@comp.nus.edu.sg}
\affiliation{%
  \institution{National University of Singapore}
}
\author{Sai Vineeth Kalluru Srinivas}
\email{sai.vineeth.kalluru.srinivas@west.cmu.edu}
\affiliation{%
  \institution{Carnegie Mellon University}
}
\author{Faisal Alqarni}
\email{falqarni@andrew.cmu.edu}
\affiliation{%
  \institution{Carnegie Mellon University}
}
\author{Shi Su}
\email{shis@andrew.cmu.edu}
\affiliation{%
  \institution{Carnegie Mellon University}
}
\author{Shijia Pan}
\email{span24@ucmerced.edu}
\affiliation{%
  \institution{University of California Merced}
}
\author{Hae Young Noh}
\email{noh@cmu.edu}
\affiliation{%
  \institution{Carnegie Mellon University}
}
\author{Pei Zhang}
\email{peizhang@cmu.edu}
\affiliation{%
  \institution{Carnegie Mellon University}
}
\author{Marco Gruteser}
\email{gruteser@winlab.rutgers.edu}
\affiliation{%
  \institution{	Rutgers University}
}
\author{Patrick Tague}
\email{tague@cmu.edu}
\affiliation{%
  \institution{Carnegie Mellon University}
}

\begin{abstract}
The resolution of GPS measurements, especially in urban areas, is insufficient for identifying a vehicle's lane. While past works have suggested augmenting coarse GPS readings with inertial sensor information for finer localization, state-of-the-art techniques do not yield enough precision to accurately pinpoint the specific \textit{lane} a vehicle is on. This impedes the realization of many novel applications like %
fine-grained navigation that can detect unsafe or infeasible turns and road planning. %
In this work, we develop a deep LSTM neural network model \altname that determines the lane vehicles are on by periodically classifying accelerometer samples collected by vehicles as they drive in real time. %
Our key finding is that even adjacent patches of road surfaces contain characteristics that are sufficiently unique to differentiate between lanes, i.e., roads inherently exhibit differing bumps, cracks, potholes, and surface unevenness. Cars can capture this road surface information as they drive using inexpensive, easy-to-install accelerometers, that increasingly come fitted in cars and can be accessed via the CAN-bus.
We collect an aggregate of $60$~km driving data and synthesize more based on this that capture factors such as variable driving speed, vehicle suspensions, and accelerometer noise. Our formulated LSTM-based deep learning model, \altname, learns lane-specific sequences of road surface events (bumps, cracks etc.) and yields $100\%$ lane classification accuracy with $200$~meters of driving data, achieving over $90\%$ with just $100$~m (correspondingly to roughly one minute of driving). %
We design the \altname model to be practical for use in real-time lane classification and show with extensive experiments that \altname yields high classification accuracy even on smooth roads, on large multi-lane roads, and on drives with frequent lane changes. Since different road surfaces have different inherent characteristics or entropy, we excavate our neural network model and discover a mechanism to easily characterize the achievable classification accuracies in a road over various driving distances by training the model just once. We present \altname as a low-cost, easily deployable and highly accurate way to achieve fine-grained lane identification.%

\end{abstract}

\begin{document}

\maketitle

\section{Introduction}
\label{sec:intro}
Development of safe transportation infrastructure and efficient mobility management are major aspects of envisioned smart cities. Finer-grained localization of vehicles and thereby their precise lane identification enables many of these use cases. For instance, lane identification aids in safer navigation, e.g., by detecting when a prescribed turn is challenging to make due to the vehicle's current lane or proactively avoiding a known rough patch on the current lane surface~\cite{L3}. %
It also aids in traffic management and road planning~\cite{bureau_2019, fernandez2017studying}, e.g., by identifying heavily used lanes that may require more frequent maintenance or expansion. Monitoring finer per-vehicle driving patterns could enable insurance use cases such as identifying rash drivers who make frequent and unsafe lane changes. In-fact, lane identification data could augment existing auto-insurance solutions to prorate insurance cost based on driving habits inferred by sensors from OBD dongle~\cite{progressive}.

It is evident that Global Positioning System (GPS) modules that vehicles and smartphones come equipped with are unreliable for these uses. %
Readings from commercial GPS modules have errors up to hundreds of meters, especially in urban areas, due to reflections from high-rise buildings that cause multi-path interference, also known as the urban canyon effect~\cite{urbancanyon}. Even high-precision GPS modules~\cite{conker2003modeling} 
that advertise sub-meter accuracy in unobstructed clear view still exhibit these problems in urban areas with errors up to tens of meters (shown in our evaluation and previous work~\cite{carloc}). To address this, researchers have proposed to leverage off-the-shelf inertial sensors to augment these error-prone GPS readings with information on vehicle trajectory/surrounding roads~\cite{carloc,semMatch,smartLoc,L3}.
However, the resolutions offered by state-of-the-art solutions are still insufficient for precise lane identification (see Section~\ref{sec:related}).
\begin{figure}[!t]
    \centering
    \includegraphics[width=0.6\columnwidth]{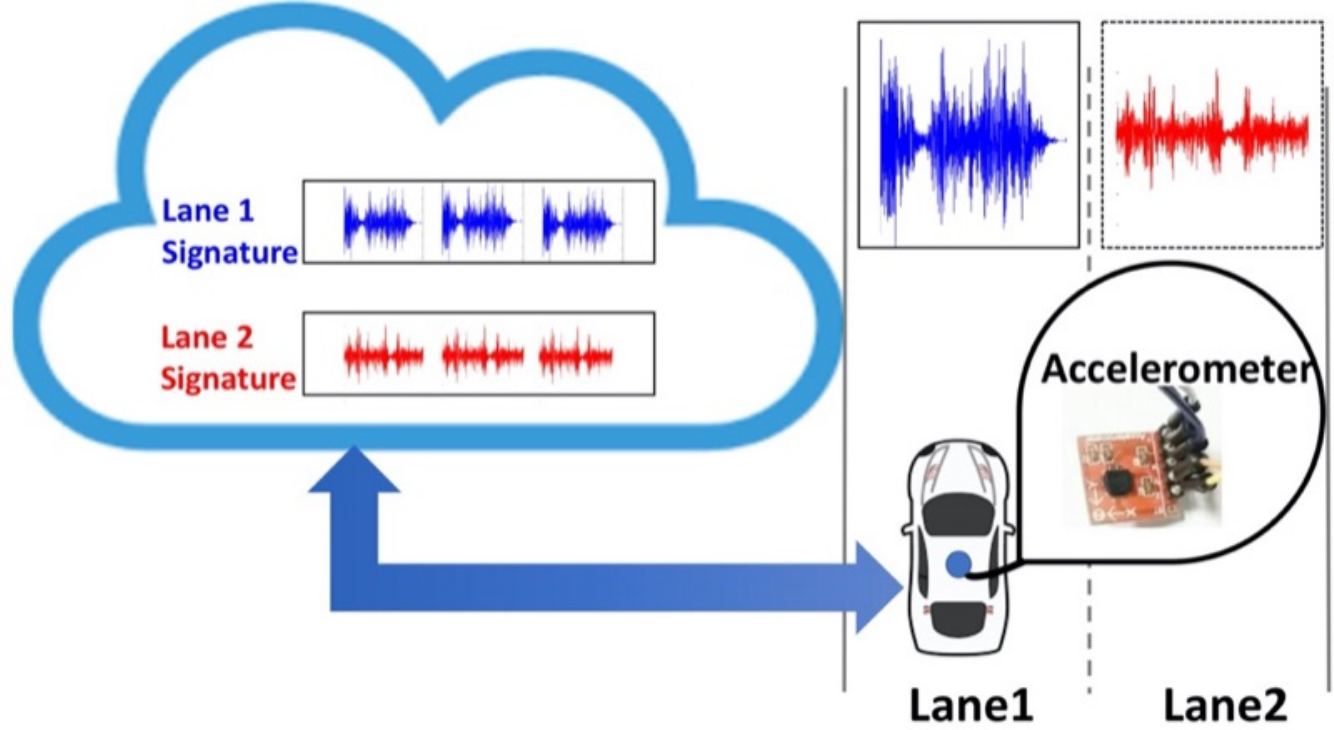}
    \caption{\altname is based on the observation that road surface characteristics can be captured by a vehicle's accelerometer and used to differentiate between adjacent lanes.}%
    \label{fig:firstOverview}
\end{figure}
Further, while camera-based solutions~\cite{calderoni2014deploying, chen2015invisible, neven2018towards} can facilitate vehicular tracking, they require widespread deployment of specialized hardware and incur privacy as well as certain functionality concerns. %

In this work, we achieve fine-grained vehicular tracking %
with driving data collected from off-the-shelf accelerometers that are widely built into cars/smartphones.
Our key finding %
is that \textbf{subtle road characteristics like bumps and cracks can be captured by inexpensive accelerometers and %
used to create ``lane signatures'' that are adequately distinguishable even between adjacent lanes}, as illustrated in Figure~\ref{fig:firstOverview}. Indeed, data from a small subset of vehicles are equipped with specialized hardware to map their collected accelerometer data to the lane within the road that it was collected on is sufficient to enable the construction of these lane signatures. %
These road surface variations that lead to lane differentiation are also evident from numerous \textit{pavement condition index} (PCI) reports that different cities regularly collect for maintenance purposes~\cite{sj_pci,sf_pci,sd_pci,dc_pci}. %
We develop \textbf{\altname, a neural network model for lane identification that learns lane-specific road surface characteristics from accelerometer data collected by few camera-equipped vehicles as they drive along roads, and subsequently classifies accelerometer samples from any vehicle in real time.)}%

\begin{figure}[!t]
\centering
\subfigure[Traditional map-matching with GPS traces]{
  \includegraphics[width=0.48\columnwidth]{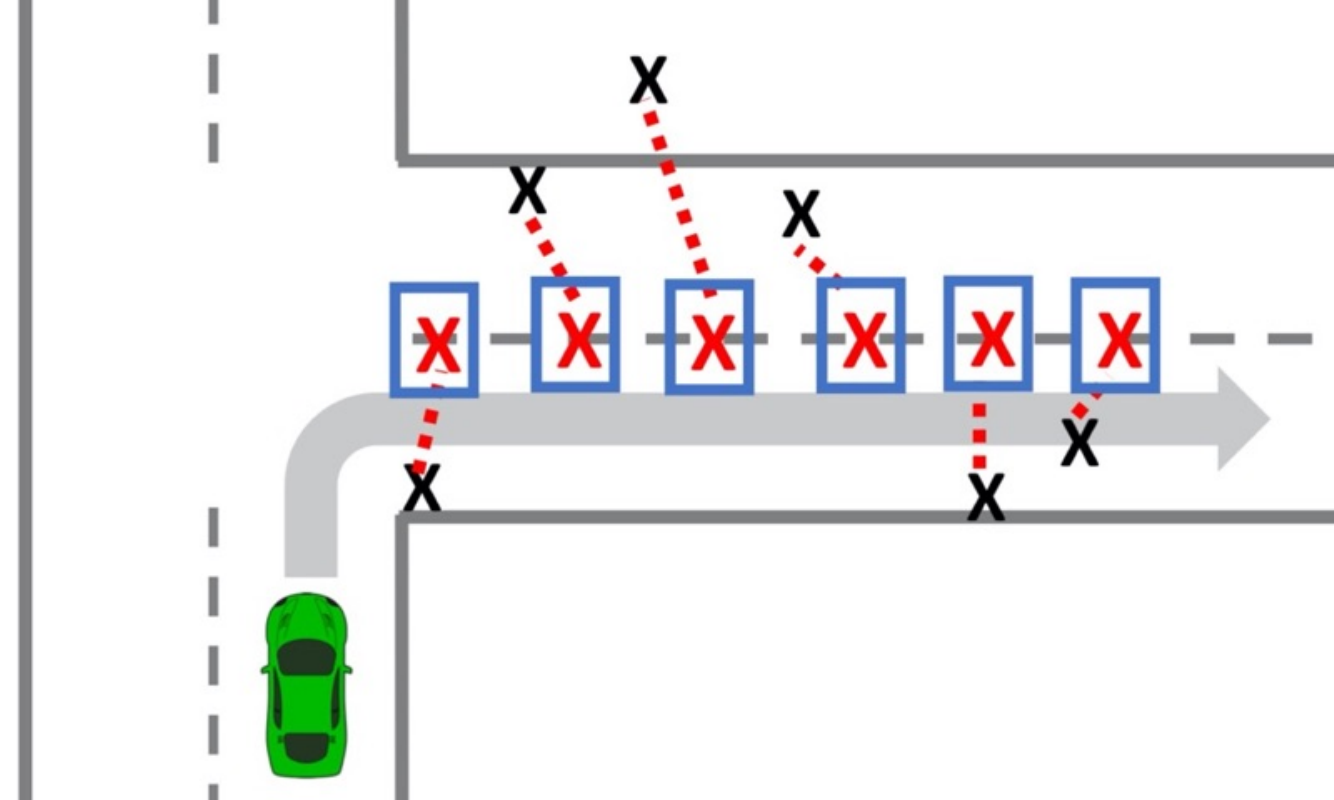} %
  \label{fig:computeCandidateRoadsA}
 }
 \subfigure[Map-matching with precise lane identification]{
  \includegraphics[width=0.48\columnwidth]{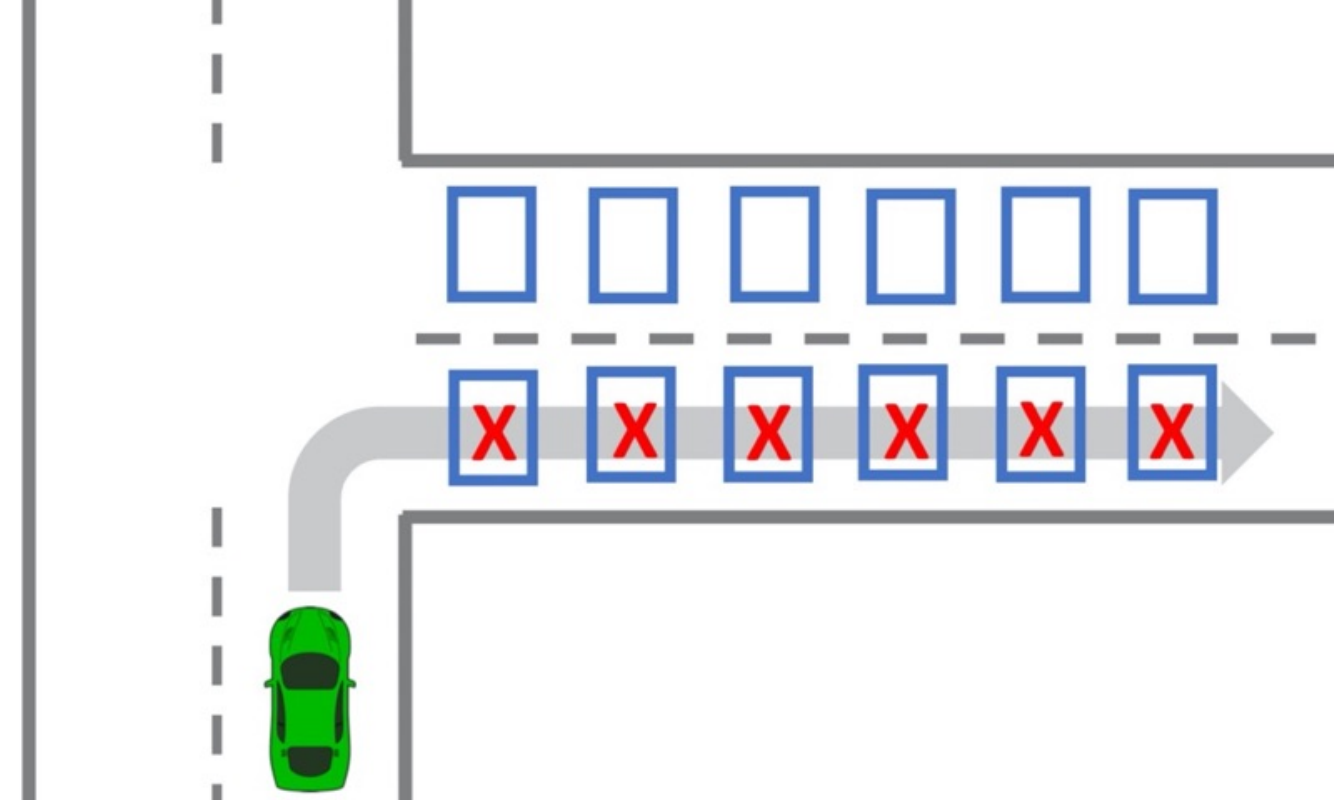} %
  \label{fig:computeCandidateRoadsB}
 }
\caption{(a) Generic map-matching is applied to the raw GPS data (shown by ``X''), and the computed path is indicated by squared ``X''. (b) By projecting the GPS data on the specific lane identified by \altname, a finer-grained path is computed.}%
\label{fig:computeCandidateRoads}
\end{figure}

Discrete lane identification %
via \altname provides a highly usable contextual property of a vehicle's position for several transportation uses, including map-matching/vehicle localization. %
With noisy GPS readings, traditional map-matching algorithms~\cite{hmmMapMatch} identify candidate roads that the vehicle may be on, which can be further refined by \altname to pinpoint the exact lane and reduce the projection space for the GPS readings to one specific lane (shown in Figure~\ref{fig:computeCandidateRoads}). With such fine-grained lane identification, navigation applications can proactively prescribe a lane change when an accident is detected in the current one father ahead and provide lane-level instructions to make upcoming turns easily accessible. Lane identification also allows infrastructure planners to identify sections of roads that are more frequently used, load balance between traffic on lanes (e.g, carpool lanes, bus lanes) and count vehicles on a per-lane basis.

We encounter several important considerations in designing \altname. First, for \altname to be in real time, it must classify accelerometer samples spanning small sections of the road that vehicles drive over in real time while trained to learn the entire surface signature of the road. Further, it is unclear what the road surface impact is on the expected lane classification accuracy or how to characterize these variations across roads. For instance, roads with smoother surfaces may presumably result in low classification accuracy due to a lack of sufficiently unique information to distinguish between lanes of smaller road sections. In other words, for \altname to be practically feasible, we need a way to assess the inherent entropy in roads and characterize the lane classification accuracy that is achievable. We address these challenges in the following ways. \textbf{ (1) We formulate a deep Long Short Term Memory (LSTM)-based neural network architecture to model the underlying sequence of road patterns for each lane.} Our modelling technique successfully extracts lane-specific features from accelerometer driving data and yields high classification accuracy even on samples representing smaller portions of the lane from vehicles driving at different speeds. \textbf{(2) To study the effect of driving distances on the classification accuracy across different roads, we collect over $60$~km of driving data in two cities containing smooth as well as rough road sections, and show that over $90$\% classification accuracy is achieved within $100$~m of driving in a lane.} However, in practice, it is challenging to train multiple \altname models corresponding to different driving distances to characterize this for each road. \textbf{(3) We realize that the hidden states of intermediate LSTM cells in the final layer of the \altname model provide an extremely useful view into the road surface as a whole, and discover a mechanism for characterizing achievable lane identification accuracy over different distances in the entire road by training \altname just once}. In-fact, we show that the parameters from the resulting model can be reused to construct smaller models that classify on smaller driving distances \textit{without requiring any retraining}. \textbf{(4) Further, to hasten model convergence, we propose a novel loss function based on the insight that longer driving distances in a lane result in higher likelihood of correct classification.} In our experiments, the \altname model did not exceed $10$~MB in size, making it lightweight and easy to deploy for real-time use.

Second, to be useful in real-time driving, \altname must react quickly to lane changes and identify the new lane switched to. Hence model responsiveness to lane switches and timeliness of classification is important. When trained on drives that span entire lanes of the road (i.e. without any lane changes), the model learns lane-specific patterns and distinguishes between them with high accuracy. In learning from drives with lane changes, however, \altname must also model the \textit{adjacency} between the two lanes, i.e., the transition probabilities of changing from a lane to another. For instance, that a transition may occur from the $50^{th}$~m of Lane 1 to the $51^{st}$~m of Lane 2 but not to the $100^{th}$~m of Lane 2 since vehicles cannot (yet) teleport. We handle lane-change concerns by doing the following. \textbf{(1) We train the model on new drives constructed from the original ones we collected (without lane changes) wherein we emulate unrealistically frequent lane switches over the entire route, e.g. at the frequency of every $25$ or $50$ meters.} However, assessing model performance on these lane-changing drives is not straightforward since the final LSTM cell of the model then provides a prediction only on the final lane segment that was switched to. \textbf{(2) We define two new metrics, namely, \textit{window of classification opportunity} and \textit{window distance}, which, when measured at intermediate cells of the output layer, exactly capture model responsiveness to lane change events.} \textbf{(3) We propose two different techniques for computing ground truth for lane-changing drives during training time, and show that upto $97$\% classification accuracy can be achieved after just $\sim15$~m of driving in the new lane.} We also highlight a design decision to be made here that essentially trades-off model timeliness (or responsiveness to lane change events) and long-term classification accuracy. 

Third, %
training a model to generalize between speed and other undesirable sources of variation in the accelerometer samples of a lane segment is challenging. Specifically, the collected accelerometer data is inherently a function of (1) vehicle driving speed, (2) vehicle suspension/accelerometer's height from the ground, and (3) engine vibrations and inherent accelerometer noise.  In-fact, new vehicles that \altname has not been trained on may use it and it is necessary to generalize across these drive- and vehicle-specific factors inherent in the collected samples. %
With sufficient data that contains a large representation of these factors, we can potentially prevent the model from over-fitting on these parameters that are not intrinsic to the road surface. However, collecting this volume and variety of driving data is prohibitively time-consuming and especially unscalable for regions with less vehicular density. In-fact, ground truth collection for \altname would presumably be done primarily by vehicles equipped with front cameras (discussed further in Section~\ref{sec:disc}) which are relatively fewer and do not represent the distribution of all vehicle types/engines/suspensions. In-fact, to train neural networks, thousands, if not millions, or data points are required. This is infeasible even for practical \altname deployments since we must rely on a limited number of vehicles with the hardware to provide ground truth data (discussed further in Section~\ref{sec:disc}). \textbf{We hence propose a data synthesis mechanism to emulate different driving speeds, vehicle suspensions and accelerometer noise for the real-world data we collect, which results %
 in an exponentially larger sample set to train on (roughly $35000$~km of driving data), thereby achieving better \altname generalization.} We rigorously evaluate the effectiveness of our data synthesis mechanism and show that this improves \altname performance by roughly $40$\% on the test set, directly aiding in model generalizability as well as creating a large enough dataset to train the neural network on.

The rest of this paper is organized as follows. Section~\ref{sec:related} reviews state-of-the-art solutions for finer localization, road condition monitoring, camera-based localization and an overview of work using machine learning to process time-series datasets as we do. In Section~\ref{sec:design}, we elucidate our goals in designing \altname, the important use cases we aim to satisfy and the challenges that emerge in doing so. After providing a brief overview of our approach, we delve deeper into \altname's model design and training process in Section~\ref{sec:network}. Subsequently, Section~\ref{sec:eval} presents results from extensive evaluation of our system, and leads to interesting points of discussion in Section~\ref{sec:disc}. We conclude our work in Section~\ref{sec:conc}.

\section{Related Work}
\label{sec:related}
We summarize related work in various relevant aspects of localization and road event detection. With respect to our neural network approach, we also review popular models for human activity recognition, and highlight differences in the vehicular context.

\textbf{Inertial Sensor-based Solutions.}
\begin{figure}[!t]
    \centering
    \includegraphics[width=0.3\columnwidth]{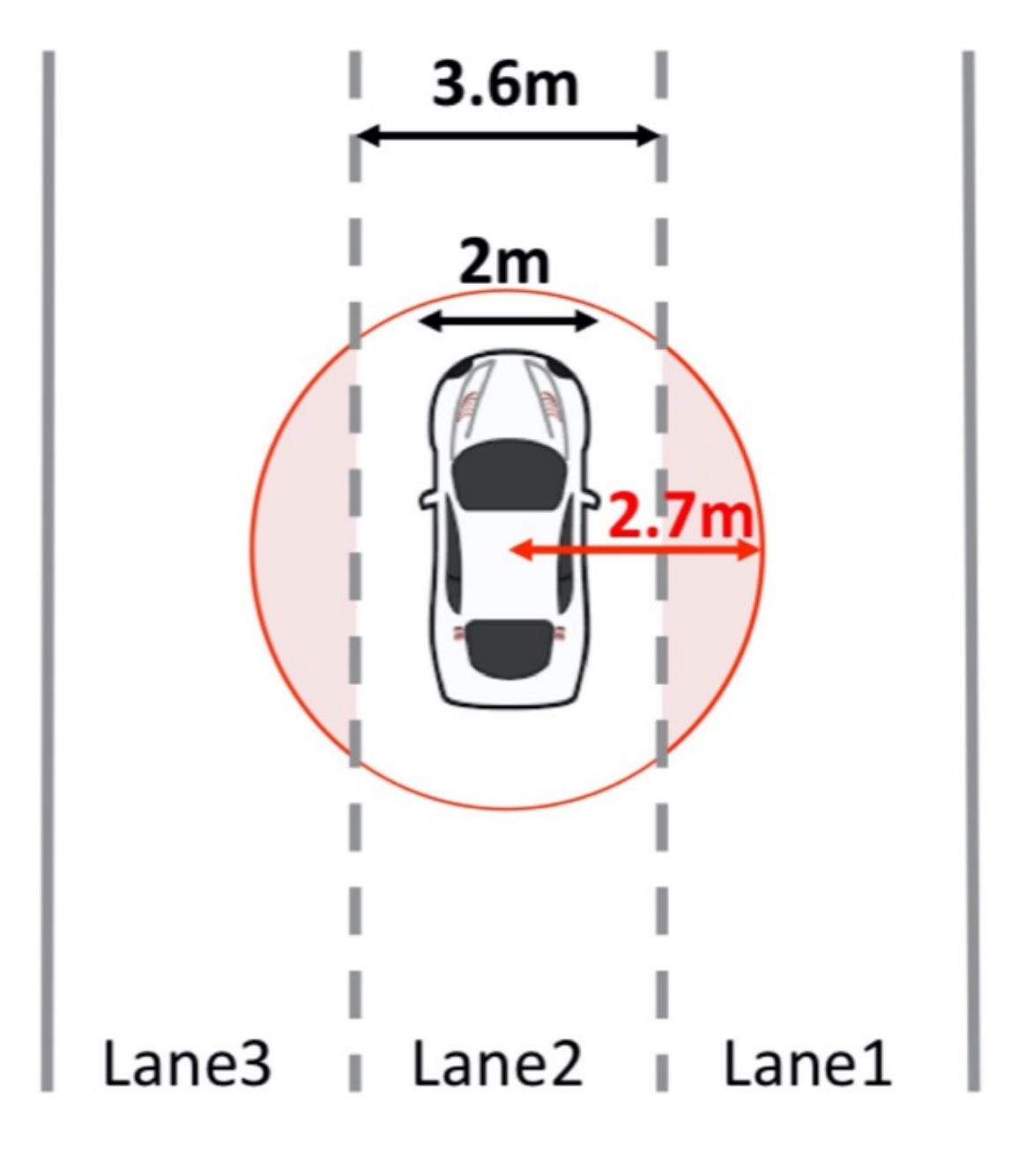}%
    \caption{We show that given typical lane and car widths of $3.6$~m and $2$~m respectively, even an optimistic GPS error bound of $2.7$~m is insufficient to accurately identify the lane a vehicle is on. For example, the GPS error in this case results in lane ambiguity between Lane1, Lane2 and Lane3.}%
    \label{fig:carlocErrorDiagram}
\end{figure}
\textit{Carloc} uses crowd-sourced location estimates of roadway landmarks, such as potholes, stop signs, traffic lights and road corners, to estimate a vehicle's location~\cite{carloc}. \textit{Carloc} achieves a mean positioning accuracy of $2.7$~m in urban areas (worst case of $4.9$~m) which, as shown in Figure~\ref{fig:carlocErrorDiagram}, is still insufficient for lane-level granularity in localization given typical lane and car widths~\cite{laneWidth, carwidth}. While \altname uses road information as well, it (1) constructs unique lane signatures from road surface data for real-time classification rather than using sporadic road landmarks and (2) uses an LSTM-based neural network model rather than the sequential Monte Carlo method~\cite{carloc}.
\textit{LaneQuest}~\cite{lanequest} formulates a rule-based algorithm based on crowdsourced information like location of road potholes or sensed information about a car's trajectory to refine the vehicle's location estimation, while \altname learns inherent road characteristics and can handle generalizable scenarios not captured by these rules. %
Other approaches~\cite{semMatch,smartLoc} aim to decrease GPS error bounds as well, but do not yield lane-level granularity or require high-end sensors~\cite{levinson2010robust}. Recent work~\cite{L3} attempts to identify the lane of a vehicle by detecting lane changes and using that to inform a probabilistic Gaussian model of the vehicle's lane. This model is, however, applicable primarily to highways. %
\altname's novelty lies in the insight that road surfaces exhibit naturally occurring fundamental characteristics that, as found in our experimentation, are distinct enough to differentiate between adjacent lanes over very short driving distances. This lends \altname more generalizable than the model proposed by Zhichen et al.~\cite{L3}. On a similar note, Chen et al.~\cite{chen2010probabilistic} propose to use Gausian Mixture Models to model crowd-sourced GPS traces and identify the lane of a vehicle. While their method yields highly accurate lane counting, \altname results in accurate lane identification for vehicles in real time. %
Fernandez et al.~\cite{fernandez2017studying} capture Wifi and Bluetooth signals of personal devices for vehicle tracking. While their solution is shown to be effective, it  requires deployment of specialized hardware in roads and does not immediately yield lane classification. 

Other works~\cite{pothole,el2017monitoring} examine the effectiveness of using solid-state sensors to detect road conditions and classifying road events (e.g. into bumps and potholes) without necessarily aiming for finer-grained real-time localization. \altname presents a concrete use of such information by comparing road conditions experienced by different vehicles on well-defined road segments to differentiate between lanes. This notion was first suggested by Han et al.~\cite{han2017convoy} to secure platooning vehicles. %
In this work, we develop a classification model based on the fundamental idea of using the road surface characteristics for lane identification of vehicles in real time, encountering various new challenges in the process. 

\textbf{Camera-based Solutions.}
Newer Advanced Driver Assistance Systems incorporate cameras to detect lanes either for lane departure warnings, or for semi-autonomous driving such as Tesla Autopilot~\cite{teslaautopilot,mobileye, neven2018towards}. These solutions are susceptible to detection error in various lighting conditions including limited visibility during night time and glares due to headlight and sunlight~\cite{chen2015invisible}. Faded lane markings and other sources of environmental noise introduce further unreliability in camera-based detection. Hence, even for cars equipped with the camera-based solutions, \altname can work in complement to achieve better localization and extend the benefits of the data collection from these specialized vehicles to the larger population of ones without front cameras. Camera solutions~\cite{calderoni2014deploying} have also been used to enable other transportation use cases like traffic monitoring and vehicle counting; \altname enables these use cases without requiring widespread deployment of specialized infrastructure by learning from the ground truth provided by these few specialized vehicles that correlates accelerometer samples from the lane they were collected in.

\textbf{Machine Learning on Accelerometer Data.} %
Our neural network approach to lane classification with time-series data has parallels in HAR.  Researchers have widely proposed~\cite{zeng2014convolutional,cho2018divide,jiang2015human,ronao2016human} 1-dimensional convolutional neural network (CNN) architectures to classify accelerometer and other time-series data from mobile sensors for human activity recognition. However, an activity like walking or jogging consists of repetitive patterns which CNNs successfully detect, but we have no reason to expect any repeating patterns in road surface characteristics. Other domains such as speech recognition and natural language processing (NLP) do work with largely non-repetitive time-series data as we do. Although speech signal processing is not perfectly applicable for lanes (for instance, Mel or MFCC~\cite{muda2010voice} techniques specific to human auditory perception lend no meaning in our context), we are inspired by widely used LSTM architectures~\cite{han2017capio, sak2014long}. %
 LSTM networks show sequence learning ability, enabling the model to, in our case, stitch together short intervals of surface patterns to learn about long sections of the lane. To design \altname, we take inspiration from tasks like sentence completion~\cite{ghosh2016contextual} %
 wherein each \textit{cell} of the LSTM layer (see Section~\ref{sec:design}) corresponds to one word of the input sentence to be completed. %
 While the lane classification problem does not exactly fit into any of these domains, we are inspired by this literature. %

\section{Lane Identification Use-Cases, Requirements and Approach}
\label{sec:design}

We first explain key usage scenarios for lane identification that drive \altname design and then describe our high-level approach to capture these requirements. In Section~\ref{sec:network}, we delve deeper into the neural network architecture.%

\subsection{Requirements and Challenges}
\label{sec:process}
The goal of \altname is to enable periodic lane identification for vehicles as they drive, by utilizing readily available infrastructure. We use the data collected by cars' accelerometers, \textit{specifically the z-axis that is normal to the ground,} to extract road information like cracks and surface unevenness. %
In designing \altname to be practically usable,  %
 we have the following important considerations.

\begin{figure}[!t]
\centering
\subfigure[Depicting variance in road surface conditions]{
  \includegraphics[width=0.4\columnwidth]{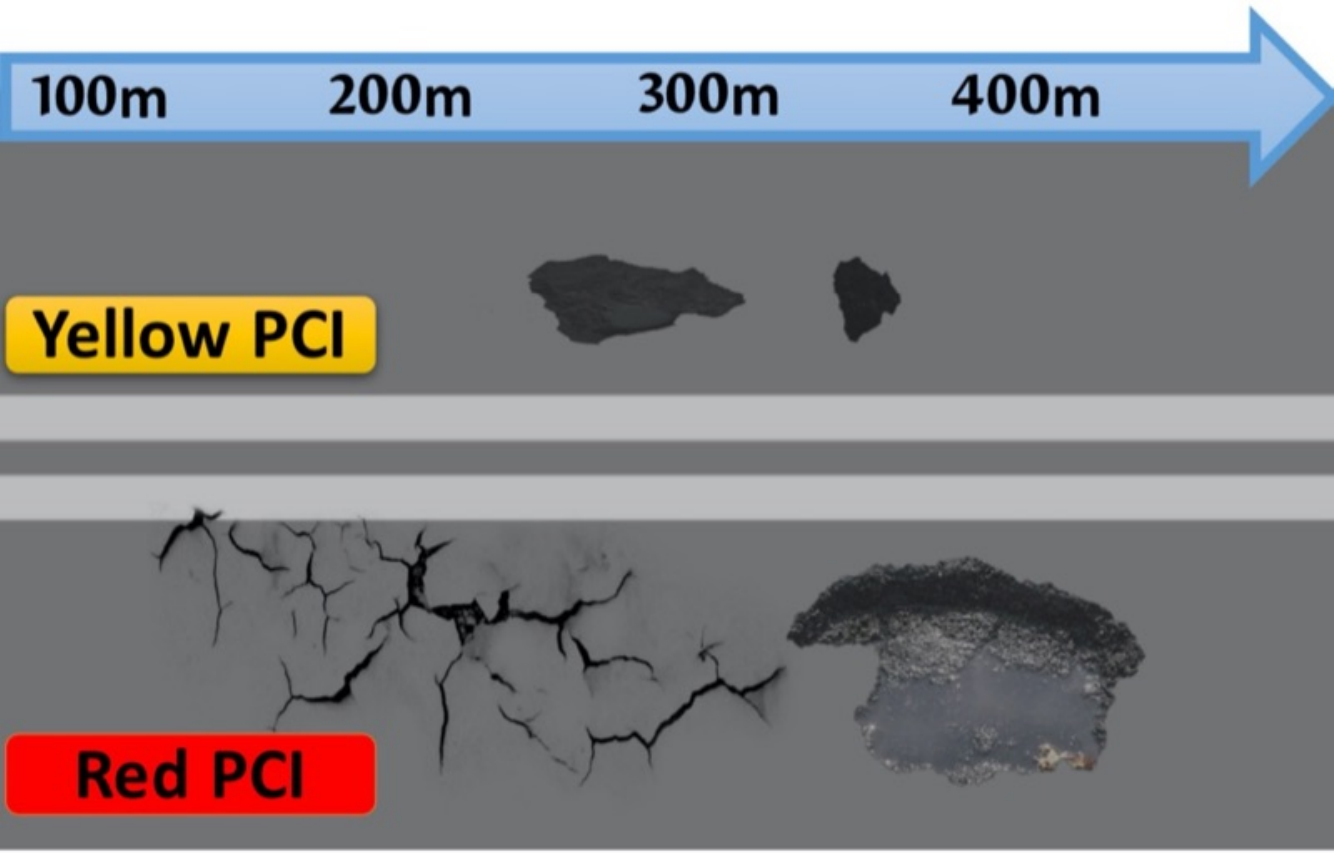} %
  \label{fig:EntropySketch}
 }
 \subfigure[Effect of vehicle speed on the accelerometer data]{
  \includegraphics[width=0.55\columnwidth]{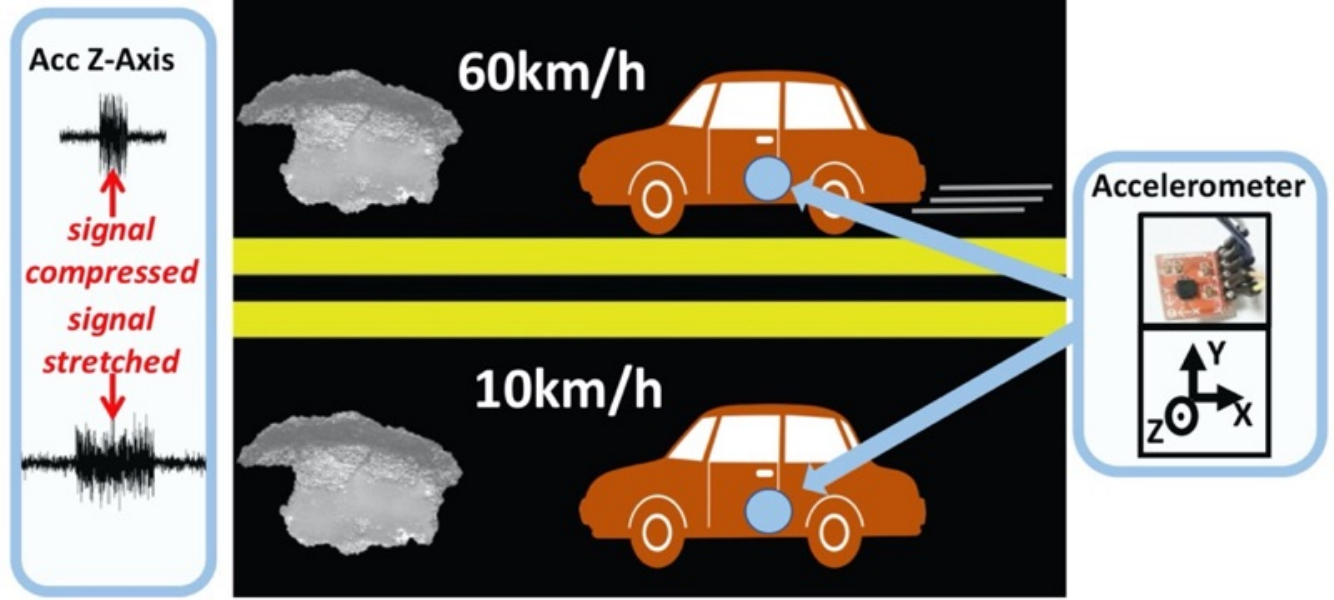} %
  \label{fig:TimeWarpingSketch}
 }
\caption{(a) There is inherent variance in road surface conditions which results in increasing lane surface information over increasing distances. (b) %
The z-axis accelerometer readings are affected by the driving speed, as shown in this illustration of two vehicles hitting a pothole at different speeds.}%
\label{fig:SignalSketch}
\end{figure}

\textbf{Learning Entire Road Signatures and Classifying Sub-Signatures.} \altname must learn the surface characteristics of well-defined road sections that presumably may span several blocks (e.g. in cities) or kilometers (e.g., between two exits on a highway). For practical uses, however, \altname must classify data from smaller sub-sections of this road as cars drive along on the road section in real time.
Therefore, while learning the lane signatures of an entire well-defined road, the model classifies portions of varying length of this road. Popular approaches to machine learning on time-series sensor data %
involve CNNs, wherein an entire discrete sample is required both for training and testing, which is not applicable here. We seek to learn the sequence of road events like bumps and cracks, as well as their amplitudes and distance in-between, for an entire road section such that even smaller portions can be distinguished.%

\textbf{Characterizing the Amount of Distinguishing Surface Information in a Road}. %
It is unclear how we can assess or characterize the unique information or ``entropy'' in a road. %
Intuitively, the longer the distance traveled on a lane, larger the amount of information captured about the road surface by the accelerometer and hence higher likelihood of correct classification. However, the smaller the minimum distance required for accurate classification in a road, the more real-time \altname's functionality is. Figure~\ref{fig:EntropySketch} illustrates such possible variance in road surface information. %
If, for instance, vehicles provide new accelerometer samples for an updated classification as frequently as once per second (e.g. like GPS refresh rates~\cite{salih2013suitability}), 
roughly $11$ to $33$~m of new driving data is essentially conveyed (given typical speeds between $38$~km/h and $120$~km/h). The amount of unique surface characteristics in the road directly impacts classification performance on this data, but it is unclear how to quantify this intrinsic entropy in different roads. For instance, in extremely smooth roads, it may simply be infeasible to distinguish between adjacent lanes with just $30$~m of driving data. This also directly impacts model responsiveness to a lane change event; on roads with lesser distinguishing information per meter, new lanes that are switched to may take longer to be detected.

\textbf{Practical Training Data Limitations.} For \altname to be easily extendable to new roads, ground truth about lane-specific road surface characteristics must be easy to acquire. The growing support for cameras in self driving cars~\cite{applecornell, nvidiablog} %
helps in correlating the accelerometer data that they collect during driving with the lane it was collected on. %
However, especially for longer road sections, it is unlikely that cars would stay on the same lane for the entire road. %
It is hence desirable to be able to learn from training samples that represent some portion a lane on the full road section of interest. Further, in practice, cars of different models will measure the road surface differently based on the suspension, accelerometer quality, etc. Different categories of cars like SUV, sedan and hatchback further have different \textit{ground clearances}~\cite{groundclearance} %
which also affect the vertical displacements of vehicles on bumps and cracks, as measured by accelerometers' z-axis. Driving speed variation is especially an important consideration, depicted in Figure~\ref{fig:TimeWarpingSketch}. In the real-world driving data we collect, we observe that the accelerometer measurements contain artifacts of these factors. \altname must generalize between these factors that inherently influence the measured data so that the ``true'' road surface pattern of each lane is learnt. We solve these learning and data challenges using a variety of techniques. We subsequently provide an overview of our approach in Sections~\ref{sec:lstm} and~\ref{sec:dataaug} and elaborate further in Section~\ref{sec:network}.

\subsection{Sequence Learning with LSTM Layers}
\label{sec:lstm}
We solve many of these challenges by using a Long Short Term Memory (LSTM)-powered neural network model for lane classification. LSTM networks, first proposed by Hochreiter et al.~\cite{hochreiter1997long}, have found considerable success in sequence learning problems wherein functions rely on a sequence of prior inputs rather than just the current one, requiring contextual understanding and information persistence of the input.

\begin{figure}[!t]
\centering
  \includegraphics[width=0.6\columnwidth]{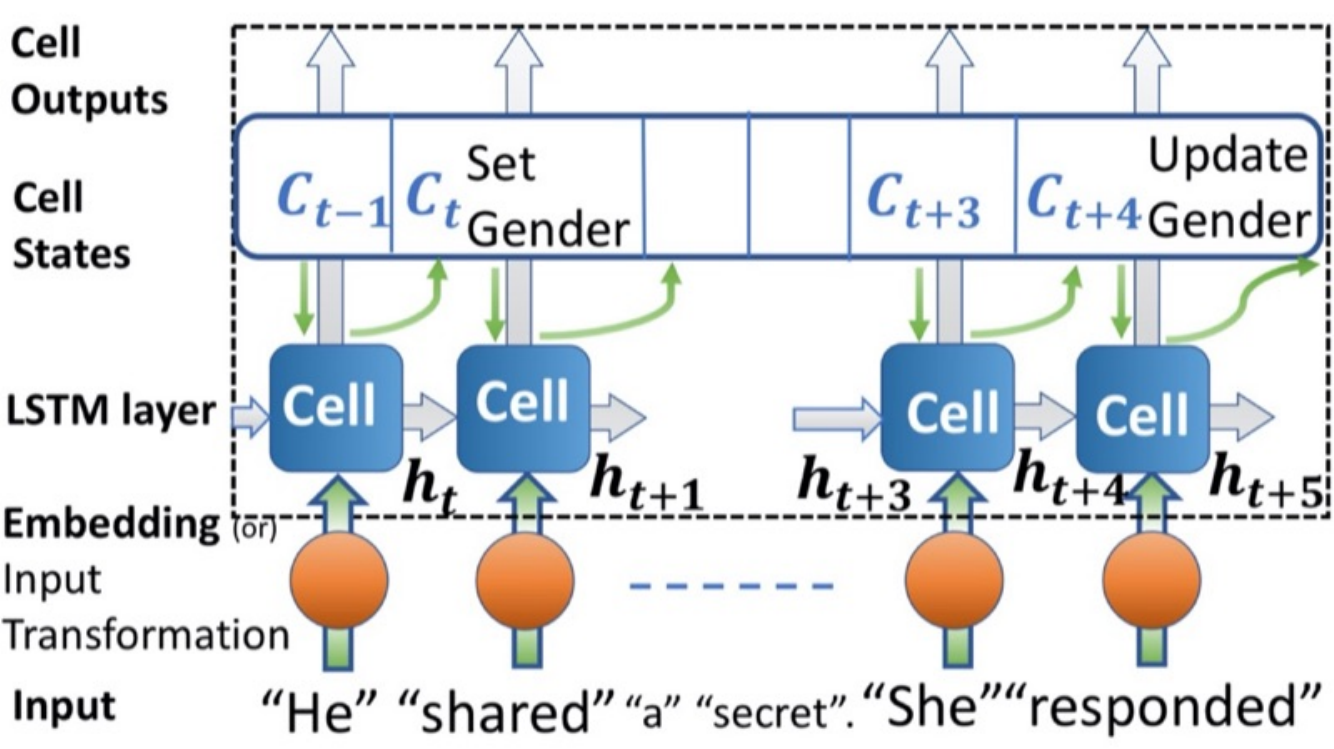} %
\caption{In an LSTM network, long-term context is captured via cell state, which is updated and propagated by each cell. Each cell computes its hidden state (also referred to as its output) based on its input, the previous cell's hidden state and the cell state.}
\label{fig:lstmcell}
\end{figure}

\textbf{Overview of LSTM networks.} LSTMs~\cite{hochreiter1997long} are a special case of Recurrent Neural Networks~\cite{sherstinsky2018fundamentals}, which contain memory cells that feed network activations from the previous time step (i.e., the computed hidden state) as input to influence predictions at the current time step. %
LSTMs additionally maintain cell states which act like a conveyor belt in allowing essential contextual information information to flow through all cells in the LSTM layer. Each cell $t$, based on the hidden state computed by the previous cell $h_{t-1}$ and its own input $i_{t}$, updates the relevant context maintained in the cell state. For instance, in a sentence completion task (for which LSTMs are widely applied~\cite{mirowski2015dependency}), a later cell may remove outdated gender information from the context $C_{t-1}$ if it receives ``She'' as input while the sentence had started with ``He said $\dots$'' (see Figure~\ref{fig:lstmcell}).
Along with providing the new context $C_{t}$, it also computes its output based on $C_{t}$, $h_{t-1}$ and its own input $i_{t}$ and offers this as the hidden state $h_{t}$ to the subsequent cell. In this example, the output may be ``responded'', forming the phrase ``She responded''. For a complete example, see~\cite{understandinglstm} and for mathematical formulations of LSTMs, see Sherstinsky et al.~\cite{sherstinsky2018fundamentals}.

\textbf{LSTMs for Lane Classification.} For lane identification, we draw parallels to other contextual tasks that use LSTMs like sentence completion, speech and digital handwriting recognition~\cite{carbune2019fast}. %
 The incremental updates as a car drives (i.e., the incremental time-series accelerometer data collected) can be considered as the current input to be classified on (i.e., the input to the last LSTM cell), while the recent history of data recorded by the car as it drove is the \textit{relevant context} (i.e., the input to the preceding cells). This framing of the lane classification problem addresses multiple concerns discussed previously. Since there is a strictly sequential relation between progressive lane segments, (e.g., a vehicle has to traverse the first $100$~m of a road to get to the next $100$~m) we can meaningfully consider the road patterns seen by a vehicle in the recent past to classify the newer samples collected. %
A few meters traveled by a car in $1$ second is not likely to contain sufficient information to distinguish between lanes on that road. In conjunction with data from several previous seconds, however, lane classification is significantly more feasible. The sequence learning enabled by LSTM also allows us to train on samples representing incremental sub-sections of the lane. %
That is, we can train the network on driving data spanning short sub-sections of the lane (as long as the entire lane is accounted for approximately uniformly in the training set), and hence classify in real time on these shorter sample lengths that the vehicle traverses rather than on the entire lane. %
This leads to easier ground truth collection for the model and allows for practical real-time use. 

In fact, our LSTM-based \altname model also facilitates characterizing the achievable classification accuracy for different driving distances in a road, making it easy to deploy widely. We elaborate on the \altname architecture and training/testing process in detail in Section~\ref{sec:network}.

\subsection{Data Augmentation for Generalization}
\label{sec:dataaug}
We now address the data challenges encountered in training \altname. %
To become agnostic to vehicle-specific factors latent in the measured accelerometer data of a lane, we can train \altname on a dataset large enough to capture all possible vehicle differences. However, this is infeasible for a variety of reasons, including impracticality of gathering data every time a new car model is produced. Indeed, data collection is a significant challenge in training neural networks for many real-world tasks. %
We propose a mechanism for augmenting our limited dataset with synthesized drives before training the \altname model. Our techniques induce variations in the original driving data that capture vehicle- as well as drive- specific factors as described below.

\textbf{Scaling.} To generalize between ground clearance variations in cars that affect the amplitudes of the accelerometer readings, we synthesize new drives from the dataset by multiplying each drive by the absolute value of a number randomly picked from a Gaussian distribution of unit mean and deviation upto 70\% of the maximum measured magnitude. %
This aids \altname in learning the pattern of bumps and relative amplitude between consecutive bumps/road events rather than memorizing a vehicle- or drive-specific magnitude.

\textbf{Jittering.} Car-specific engine vibration pattern (i.e., vibration frequency) also change with the shaft rotation rate, adding noise to the accelerometer data. %
Further, accelerometers themselves have differing inherent noise.  %
We therefore synthesize new drives by adding Gaussian noise of zero mean and differing standard deviations up to 10\% of the maximum measured amplitude to induce such engine jitter on the original and scaled samples. %

\begin{figure}[!t]
\centering
\includegraphics[width=0.45\columnwidth, trim={0 .1cm 0 0cm}, clip]{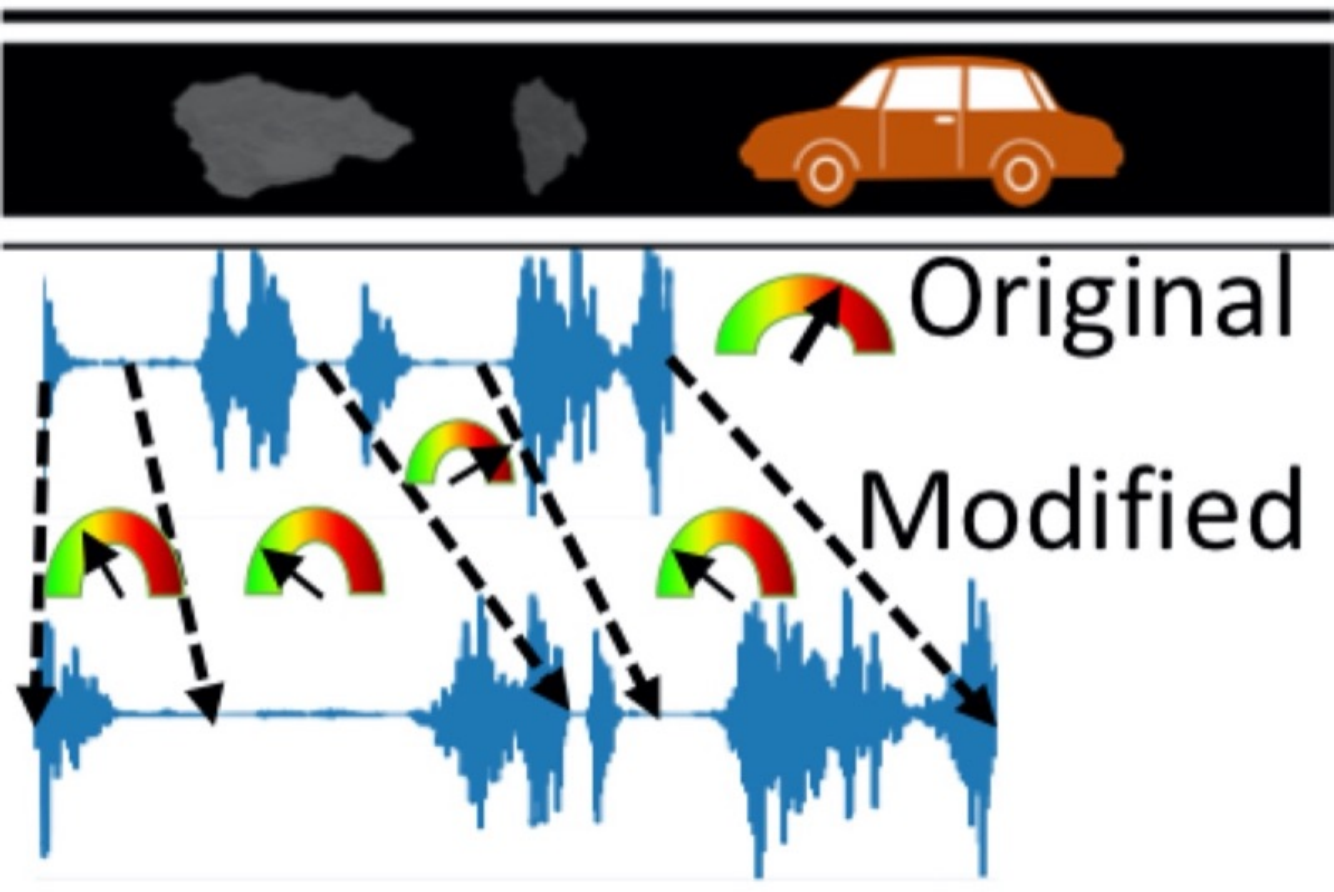} %
\caption{In synthesizing a time-warped signal, we emulate different driving speeds that result in different sections of the original signal being stretched or compressed.}
\label{fig:TimeWarpingAugmentationSketch}
\end{figure}

\textbf{Time Warping.} %
As Figure~\ref{fig:TimeWarpingSketch} shows, the accelerometer measurements for a given road segment depend on the driving speed. Hence, each drive provides a speed-dependent view of the lane while \altname must learn the ``true'' sequence of road surface events and the intervals between them.  %
We synthesize data samples to emulate vehicles driving at different speeds to provide this needed generalization. For each drive in the dataset (original as well as jittered and scaled)%
, we first approximate an interpolation function $y = f(x)$, where $x$ is the time step or sample count and $y$ is the accelerometer magnitude. We then randomly choose different subsections of this drive, in sequence, and emulate a lower or higher driving speed for that subsection by sampling from $f$ at a higher or lower frequency respectively. Hence, the synthesized drive undergoes potentially multiple accelerations and decelerations over the same distance as the original one, as shown in Figure~\ref{fig:TimeWarpingAugmentationSketch}. %
For each speed change, the new speed is kept within 20\% of the original to avoid erratic or atypical speed changes. %

With these data augmentation techniques, we capture vehicle and drive characteristics which are otherwise prohibitively time-consuming to collect manually and make our dataset large enough for neural network training.

\section{\altname Design}
\label{sec:network}
We now introduce \altname, a deep LSTM neural network architecture for lane classification. 
We first explain the construction of input sequences that the model trains/tests on, as this substantially informs the network architecture. We subsequently detail the neural network model and various design choices made.

\subsection{Neural Network Input}
\label{sec:input}

 We first segment each drive spanning a lane of the entire well-defined road (with specific start/end locations) into smaller sub-drives of length $\ell$ and choose a stride (or sliding window) of length $s$. For instance, a drive of $800$ samples with $\ell = 100$ and $s = 50$ would result in $15$ sub-drives of length $100$ samples each. If the final sub-drive, i.e. $15^{\text{th}}$ in this case, does not have the requisite $\ell$ samples, we pad it with zeros as required.

Each training sample that the model trains on corresponds to a sub-drive, and sub-drives from multiple drives are randomly shuffled before forming a batch to train on. By training on these sub-drives that represent different sub-sections of the road to learn, we allow the model to generalize between specific start/end points and instead learn the sequence of road surface characteristics spanning the entire lane. Further, in realistic deployments, ground truth may be collected from cars driving over a small portion of a lane within a road section before changing roads or changing lanes. %
In this case, as long as $\ell$ samples are collected in the road, that portion of the drive may be used for training \altname. The \textbf{``bootstrapping time''}, i.e., the amount of accelerometer samples required to get the first lane classification result from \altname when a car starts driving, is directly set by $\ell$, since $\ell$ samples are required to construct a train or test sample. However, note that once the first $\ell$ samples are collected, the vehicle may poll \altname for updated classification at even 1-sample frequency and, in that case, simply provide the 1 new sample along with the last $\ell-1$ samples collected. In Section~\ref{sec:eval}, we experimentally measure the trade-off between increasing classification accuracy with more samples and lowering $\ell$ for faster bootstrapping time.

For actually training or testing the model with a sub-drive, we segment the sub-drive further. Sub-drives are divided into \textit{sub-segments} of length $d$ and stride $m=d/2$. Hence for every sub-drive, a total of $n = (\ell - d)/m + 1$ sub-segments are generated. We experimentally observe that training on this set of sub-segments that correspond to a sub-drive, as constructed with 50\% overlap, enables \altname to generalize between driving speeds much more effectively than training directly on the sub-drive, akin to prior findings~\cite{azzouni2017long}. In the subsequent section, we present the \altname architecture and explain how these $n$ sub-segments provided in an input sample is handled. %

\subsection{Neural Network Architecture}
\begin{figure}
    \centering
    \includegraphics[width = .6\columnwidth]{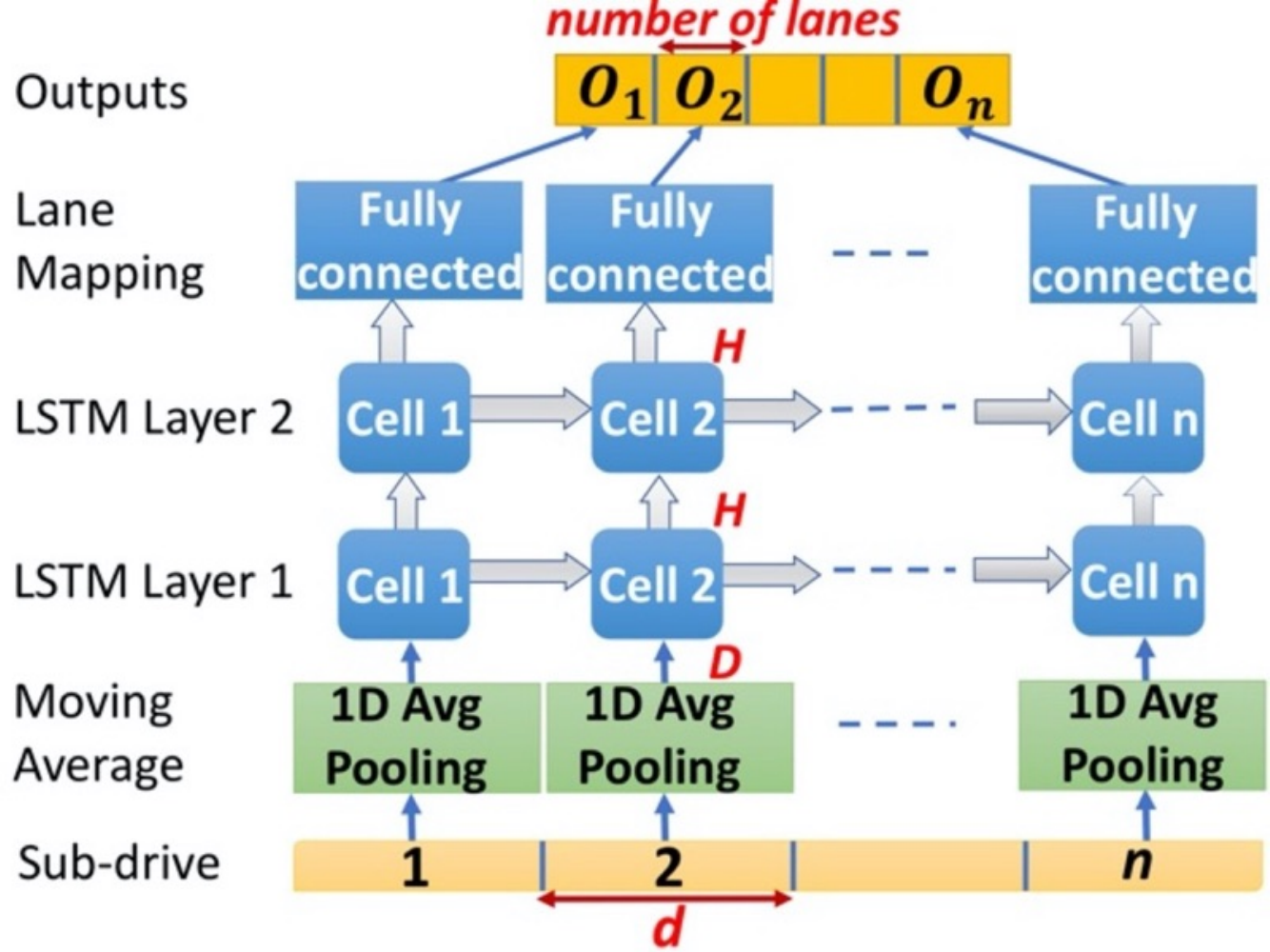}
    \caption{We illustrate \altname's deep-LSTM architecture. Trainable layers and dimension of intermediate outputs are shown in blue and red respectively.}
    \label{fig:lstm_arch}
\end{figure}

Figure~\ref{fig:lstm_arch} illustrates \altname architecture in the context of a single input (i.e., batch size $b$ = 1). The input consists of a sub-drive broken into $n$ sub-segments of length $d$ as described. %
We first use an average pooling layer %
to apply a sliding window average to each sub-segment, thereby reducing the number of samples while retaining much of the salient road surface information.
We denote the length of each sub-segment after applying average pooling as $D$. %
Each sub-segment of length $D$ feeds into a corresponding LSTM cell in the first LSTM layer of the network, resulting in as many as $n$ LSTM cells in this layer. The last LSTM cell operates on the samples corresponding to  ``the latest'' time step of the provided driving data input, while the first LSTM cell operates on samples collected $n$ timesteps ago with respect to the last cell. Each cell computes a representation of size $H$ of the underlying hidden state based on its provided cell-specific input as well as the hidden state computed by the previous cell (see Section~\ref{sec:lstm} and~\cite{sherstinsky2018fundamentals} for more details on how LSTM works). This result is both the cell's output as well as the cell's hidden state that is used by the next cell. %
A second LSTM layer is stacked on top of this to further search for temporal characteristics in the output of the first LSTM layer. %

Each of the $n$ cell's outputs from the second LSTM layer is also of length $H$ and is passed through a fully connected layer %
that maps this $H$-sized input to an output of size equal to the number of lanes to classify, resulting in $O_{1}$\dots$O_{n}$. We apply the softmax function~\cite{nwankpa2018activation} to each output vector $O_{i}$ to interpret the values within as the probability of the input sample belonging to the corresponding lane. %
The output corresponding to the final LSTM cell, $O_{n}$, is considered as the network prediction for the provided sub-drive. %

Note that our LSTM layers are stateless, i.e., through careful selection of $\ell$, we provide sufficient context in a given input sequence for training such that the network need not remember internal states from previous batches. This enables \altname to be offerred as a stateless service to end-users, e.g. via REST APIs, which are generally easier to deploy and maintain than stateful ones. %

\subsection{Network Output and Loss Function}
\label{sec:loss}
For LSTM-based sequence classification or sentence completion tasks, the final cell of the last LSTM layer is typically used as the network prediction, since it incorporates knowledge from all previous cells to produce its output. Sometimes, a fully connected layer maps this last LSTM cell to the number of output classes to produce the final prediction. In either case, the classification loss then computed is a direct function only of the final LSTM cell's output~\cite{greff2016lstm}. %
For \altname, we instead attach a fully connected layer to each cell $i$ of the final LSTM layer, as described above, which maps to one of the lanes as output $O_{i}$. Denoting the target or ground truth for the input sub-drive as $O^*$, we back-propagate the loss $\mathcal{L}(\{O_i\},O^*)$ defined as 
\begin{equation}
    \mathcal{L}(\{O_i\},O^*) = \sum_{i=1}^n \frac{2i}{n(n+1)}\text{CELoss}(O_{i}, O^*)
\label{eq:lossEq}
\end{equation}
\block{ %
\begin{align}
    \begin{split}
        \sum_{i = 1}^{n} &w_{i, \text{norm}}*\text{CELoss}(O_{i}, t) \text{, where}, \\
        w_{i} &= \frac{i}{n}, \\
        w_{i, \text{norm}} &= \frac{w_{i}}{\sum_{i = 1}^{n} w_{i}} 
    \end{split}
\end{align}
}
In essence, we weight the cross-entropy loss $\text{CELoss}(O_{i}, O^*)$ from each classification result $O_{i}$ by cell $i$'s distance from the final cell (weights normalized to sum to $1$) so that classification inaccuracy in later cells is penalized more than earlier ones. The choice of weights in this loss function guides the loss surface generated and hence the parameter exploration. The proposed cell-order based weighting function (as opposed to, for instance, uniform weighting) directly captures the intuition that cells in earlier stages of the LSTM operate on less aggregate information than subsequent ones. %
When the input, representing a sub-drive of length $\ell$, corresponds to exactly one lane that the sub-drive was driven on, our loss function captures the contextual notion of increasing information about the lane surface as the vehicle drives, and hence increasing likelihood of lane differentiation. %
 For instance, the first cell, operating on just one $D$-sized input, may be unable to distinguish between two lanes. However, subsequent cells like $4$, $7$, and $9$, operating not just on their own input of length $D$, but also on information from previous cells, are expected to yield higher classification accuracy. As we see in Section~\ref{sec:eval}, employing this loss function even yields intermediate cells optimized for classification accuracy, thereby allowing us to easily characterize the entropy of new roads, and faster model convergence.

\section{Evaluation}%
\label{sec:eval}
We now evaluate \altname's performance in multiple ways. First, we test the effectiveness of our data augmentation mechanism as the synthesized drives form the bulk of the dataset that \altname is trained and tested on. We then test the weighted loss function that we propose in (\ref{eq:lossEq}) for training on drives spanning entire lanes of the road and illustrate the faster molde convergence that results. After this, we test \altname's ability to distinguish between two adjacent lanes as a function of the distance traveled in a lane, across routes in two cities. We also study the impact of overall road condition (worn out vs. smooth) on this. Next, we test the model performance over increasing number of lanes and show that \altname is able to distinguish between siz lanes with high accuracy. We then consider the challenging scenario of frequent lane switches and study \altname's classification accuracy when vehicles switch lanes as often as every ~$50$ metres (i.e. 1-4 seconds), across routes in both cities. In doing so, we also discover interpretations into the model's learning function, which is generally very challenging to do for neural networks. These interpretations provide guidance in the choice of certain key design decisions.

Finally, we present a mechanism for characterising the entire road surface's inherent entropy. Typically, for LSTM-based neural networks, the final cell's result from the last layer alone is considered as the network output (a.k.a model prediction) and the rest of the outputs discarded/unused. However, by analyzing intermediate cell outputs from the neural network's last layer, we realize that these provide an extremely useful view into the road surface structure. In fact by training the model once on driving distances that span the full route, we are able to characterize the growth in lane classification accuracy over increasing distances in the lane and thereby the inherent surface information along the road. As we show, this significantly eases \altname's training/deployment burden and makes \altname highly practical. %

\begin{figure*}[!t]
    \centering
    \subfigure[San Francisco route]{
    \includegraphics[width=0.25\textwidth]{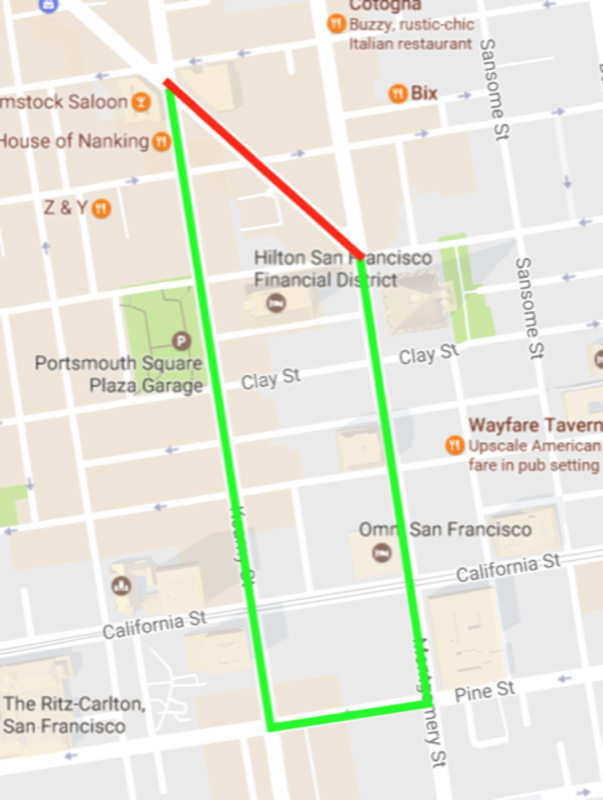}
    \label{fig:pci_png_sf}
    }
    \subfigure[San Jose route]{
    \includegraphics[width=0.25\textwidth, height=.32\columnwidth]{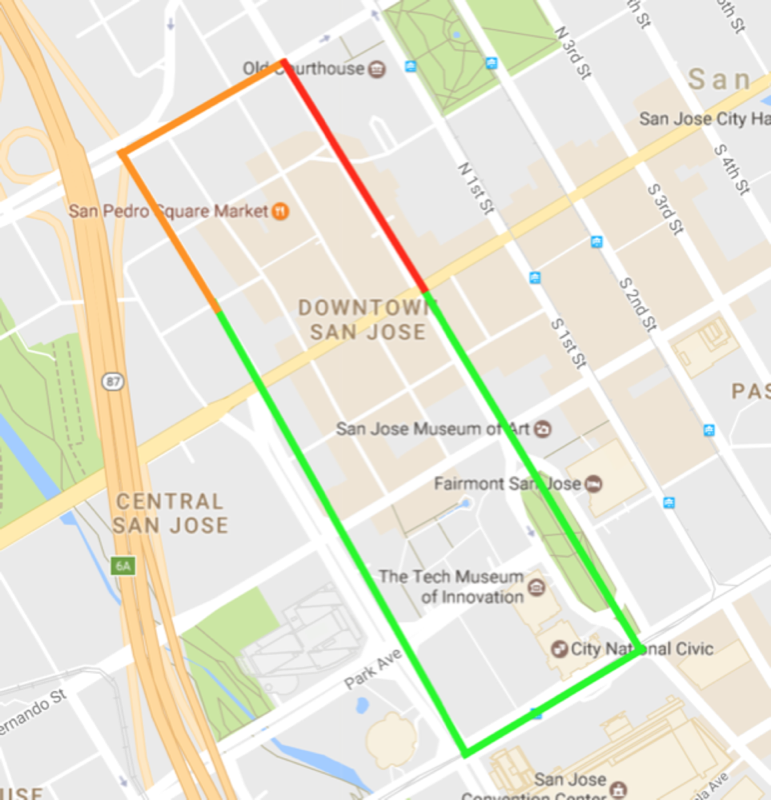}
    \label{fig:pci_png_sj}
    }
    \subfigure[GPS traces for San Francisco]{
      \includegraphics[width=0.21\textwidth]{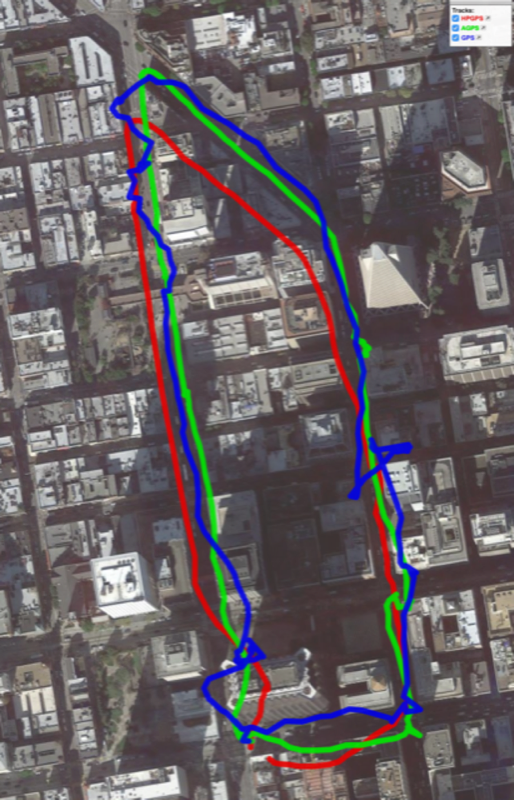}
      \label{fig:Aerial_SF}
     }
     \subfigure[GPS traces for San Jose]{
      \includegraphics[width=0.21\textwidth]{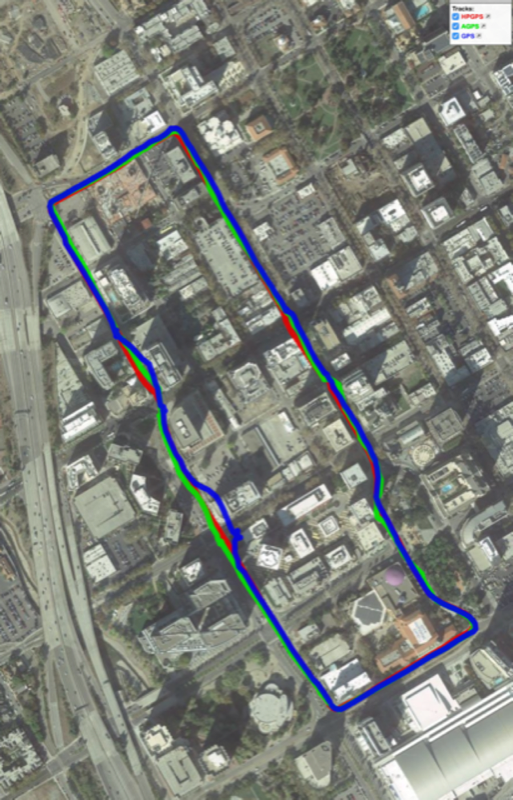}
      \label{fig:Aerial_SJ}
     }
\caption{The routes driven in San Francisco (a) and San Jose (b) capture smooth as well as rough road sections, as indicated by the colors of the PCI scores for each road (green/yellow/red corresponding to fair/moderate/poor). By overlaying GPS, AGPS and HPGPS measurements onto the map of both these routes (c, d), we see that these traces are noisy.} %
\label{fig:drivingroute}
\end{figure*}

\textbf{Setup and Appatarus.} We pick a route of length 1.2~km in downtown San Francisco (SF), shown in Figure~\ref{fig:pci_png_sf}, wherein we drive ten times each over two adjacent lanes with a 2014 Volkswagen Jetta and 2012 Subaru Impreza, yielding $40$ drives in total. We also drive over a 2.4~km route with two lanes in downtown San Jose (SJ) (Figure~\ref{fig:pci_png_sj}) with the Volkswagen, for five separate trials per lane, yielding $10$ drives. In aggregate, we collect 60~km of real-world driving data. Note that these routes span multiple intersection with stop signs and traffic lights.
We collect the following sensor readings from each drive. First, we measure accelerometer z-axis (i.e., normal to the road surface) from a +/-3g triple-axis MEMS accelerometer~\cite{adxl335} interfaced to an Arduino UNO~\cite{arduino}. The accelerometer is \textit{firmly affixed to the floor of the vehicle.} Note that most vehicles have OBD ports from where this accelerometer data can be easily acquired, which is then impervious to concerns of accelerometer orientation. In practice, accelerometer samples collected by smartphones can be used with \altname as well by applying orientaton-correcting techniques~\cite{tundo2013correcting, chen2015invisible}. The accelerometer uses a sampling rate of $6$~kHz, %
 but since \textit{we use a kernel of size $500$ and stride $50$ for downsampling the accelerometer data via the  average pooling layer in our model (Figure~\ref{fig:lstm_arch})}, \textbf{the effective sampling rate is reduced to to $111$~Hz}. For the rest of this section, references to samples are in terms of our original $6$~kHz sampling rate. The average number of samples collected in the SF and SJ drives are about $1.2$M and $1.8$M respectively; hence $100$K samples represent approximately $95 - 130$~m  of driving distance.

 For baseline comparison, we also collect a Satellite-Based Augmentation System (SBAS)  High Precision GPS (HPGPS) from U-Blox EVK-7P module (\textgreater \$200)~\cite{ublox}. We collect Assisted-GPS (AGPS) readings from a Nexus 4 GPS logger app~\cite{gpslogger}, wherein cellular information is used to augment GPS readings. Lastly, we collect GPS readings from a Nexus 4 without a SIM card.

\textbf{Training Settings.
} The raw signal from the accelerometer is first normalized and Hampel filtering is applied to remove noisy Arduino artifacts. Using the mechanism described in Section~\ref{sec:dataaug}, we synthesize additional data from our collected samples. We apply the \textit{scaling} technique to each drive approximately $10$ times, \textit{jittering} to the original as well as scaled drives $10$ times, and finally \textit{time-warping} to the original, scaled and jittered drives $\mathtt{\sim}5$ times. Hence, our SF and SJ datasets finally contain about $30000$ and $10000$ complete drives over the corresponding routes. We choose an 80/20 split of our data for training/testing sets respectively such that no original drive or any of its synthesized variants from one set appear in the other. We use a learning rate of $0.005$ with Adam optimizer, hidden dimension size $H = 300$ for both LSTM layers, and batch size $b=512$ with random shuffling. We generally set $d=50$K and $s=50$K; for $\ell<=200$K, we set $d=50$K to get atleast $1-7$ LSTM cells in the model and $s=10$K. Model is typically trained for $2-4$ epochs until the validation accuracy starts to decrease. %

\textbf{Baseline.} Figures ~\ref{fig:Aerial_SF} and~\ref{fig:Aerial_SJ} depict the three noisy baseline trajectories overlayed onto the maps of San Francisco and San Jose. GPS errors are more salient in San Francisco as expected, since the downtown location with taller buildings is more prone to the urban canyon effect~\cite{urbancanyon}. However, in neither route are we able to achieve lane-level localization with even HPGPS readings, as discussed later. 

\subsection{Effectiveness of Data Augmentation Techniques}%
\begin{figure}[!t]
    \centering
    \includegraphics[width=0.5\textwidth]{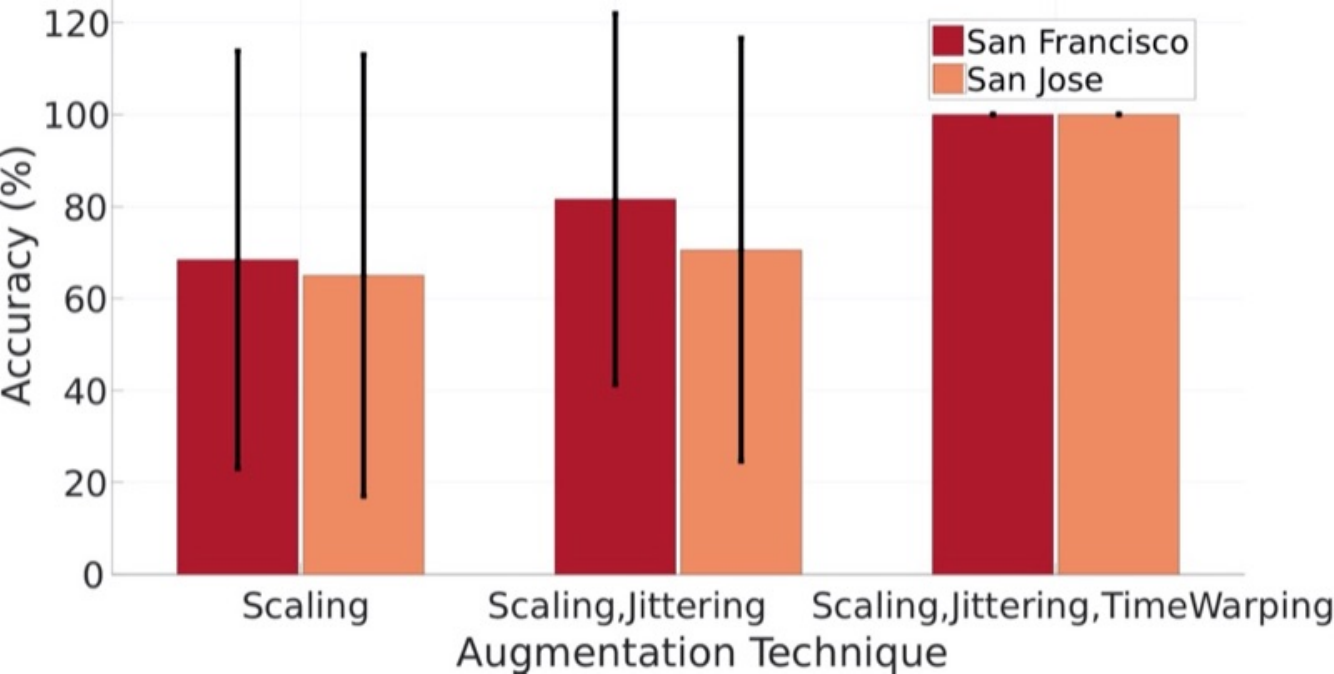}
    \caption{Across both the SF and SJ routes, \altname's mean classification accuracy and its variance on the original test set steadily improves when the training set includes jitterred and subsequently time-warped drives.}
    \label{fig:effectOfAug}
\end{figure}

We propose the data augmentation techniques in Section~\ref{sec:dataaug} to increase our dataset size in a manner that captures various real-world factors that could influence the accelerometer data collected by a vehicle, including its suspension, accelerometer noisiness and driving speeds and patterns. Having an increasing amount of these different sources of drive- and vehicle- specific variations directly aids in model generalizability. To test this, we first train the model on original and scaled drives, then on original, scaled and jitterred drives and finally on all drives (including time-warped ones).

In all cases, we train one model for SF and one for SJ. We use the segmentation procedure described in Section~\ref{sec:input} to create sub-drives of size $\ell$ from the train and test set. Presumably, smaller $\ell$ (i.e. samples spanning smaller portions of the route) may be more challenging to train on since there may not be sufficiently distinguishing lane information in small road sections. Since the impact of driving distance on model performance is evaluated in detail separately, we set $\ell$ to a large value for both the SF and SJ models to control for the impact of driving distance. Figure~\ref{fig:effectOfAug} shows the mean accuracy on the original (non-augmented drives) test set for both the SF and SJ models. As we see, the average model performance increases as the training set captures increasingly different sources of variations in the data. In-fact, the variance in the classification accuracy decreases to nearly $0$ when trained on the entire dataset containing the original and all synthesized drives, indicating the effectiveness of these techniques in increasing model generalizability, especially the time-warping procedure. Figure~\ref{fig:effectOfAug} provides a strong indication that as the number of original drives increase (to capture even more original variation in driving patterns, speeds, vehicles, etc.), the use of these techniques to synthesize additional variations from these will lend the model highly generalizable across these factors.

\subsection{Comparing Loss Functions}
\begin{figure} 
\includegraphics[width=.4\textwidth]{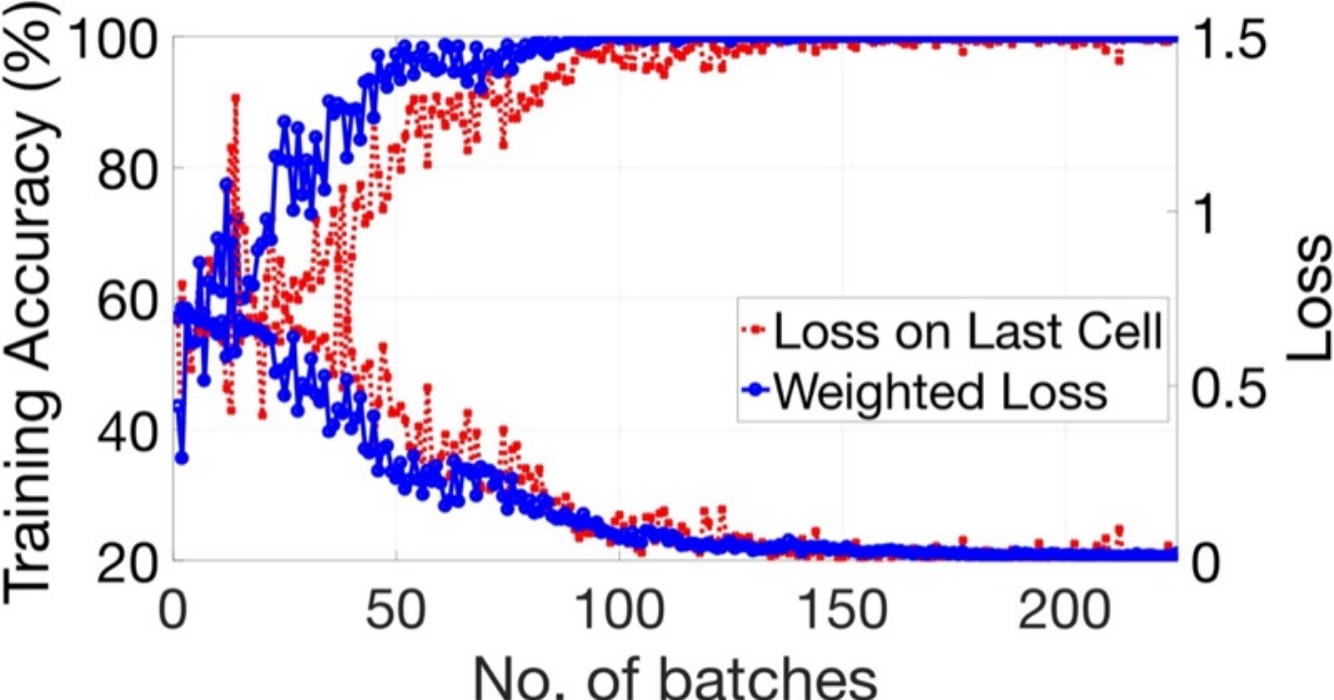}
     \caption{Our proposed weighted loss function results in steeper increase of training accuracy and faster decay of loss than computing loss on merely the last LSTM cell.}
    \label{fig:loss}
\end{figure}

As noted in Section~\ref{sec:network}, we attach a fully connected layer to each cell of the second LSTM layer, rather than merely to the final cell as typically done. We further propose a novel loss function based on this, wherein the average loss is computed across each output $O_{i}$ and weighted by the proximity of the corresponding cell to the last one. As we show, this becomes key to faster and more stable model training. Figure~\ref{fig:loss} depicts a representative instance of this effect on the SJ dataset ($\ell$ set to $1M$), as we see the training accuracy grow much faster and converge more quickly to a stable $100\%$ as compared to backpropagating the loss of the output from just the final LSTM cell. Essentially, the model benefits from the additional insight we provide via our guided loss function, namely that the classification confidence must directly increase with each LSTM cell. Since each drive spans an entire lane of the route (without lane changes),  the available information to differentiation between adjacent lanes only increases (weakly) within a sample. We later show that the \altname model, trained with this loss function on drives spanning large portions of lanes, provides crucial insights into the road surface information and characterization of achievable accuracy in a road.

\subsection{Distinguishing between Adjacent Lanes over Varying Driving Distances}
\begin{figure*}[!t]
    \centering
    \subfigure[SF -  Accuracy over varying $\ell$]{
    \includegraphics[width=0.32\textwidth]{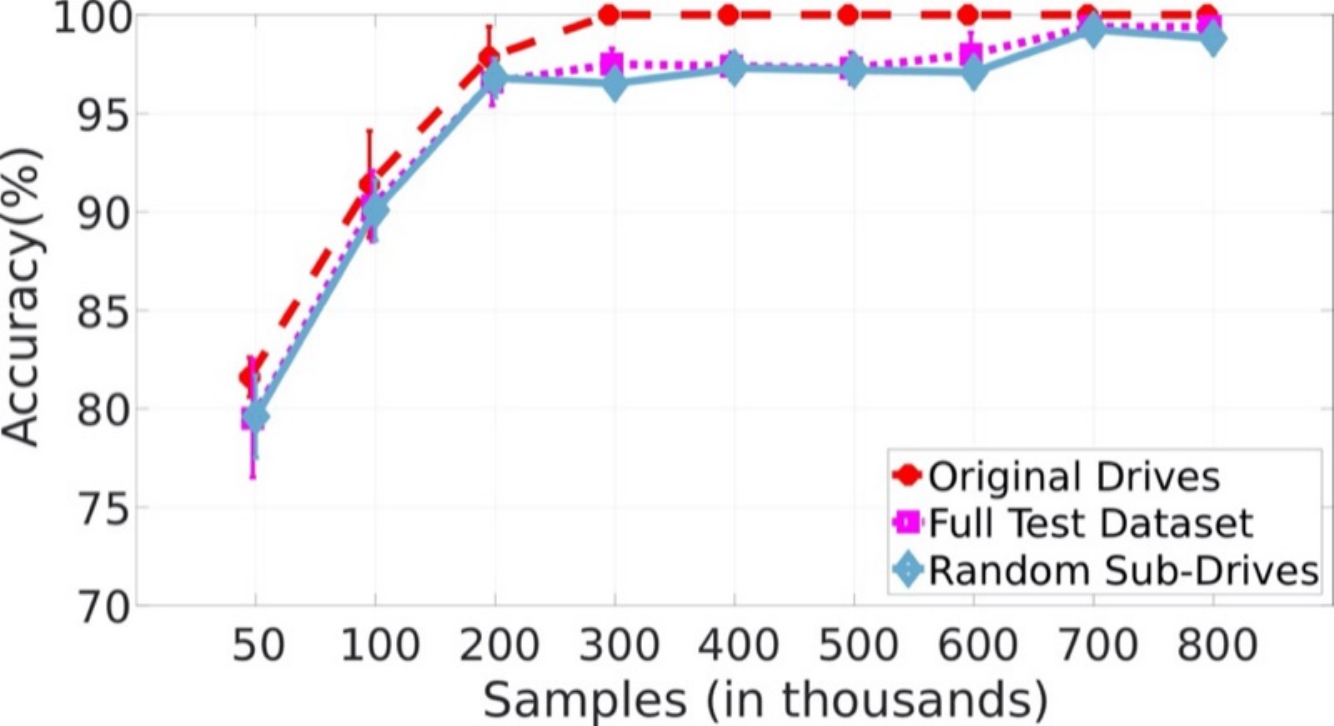}
    \label{fig:sfAcc}
    }
    \subfigure[SJ -  Accuracy over varying $\ell$]{
    \includegraphics[width=0.32\textwidth]{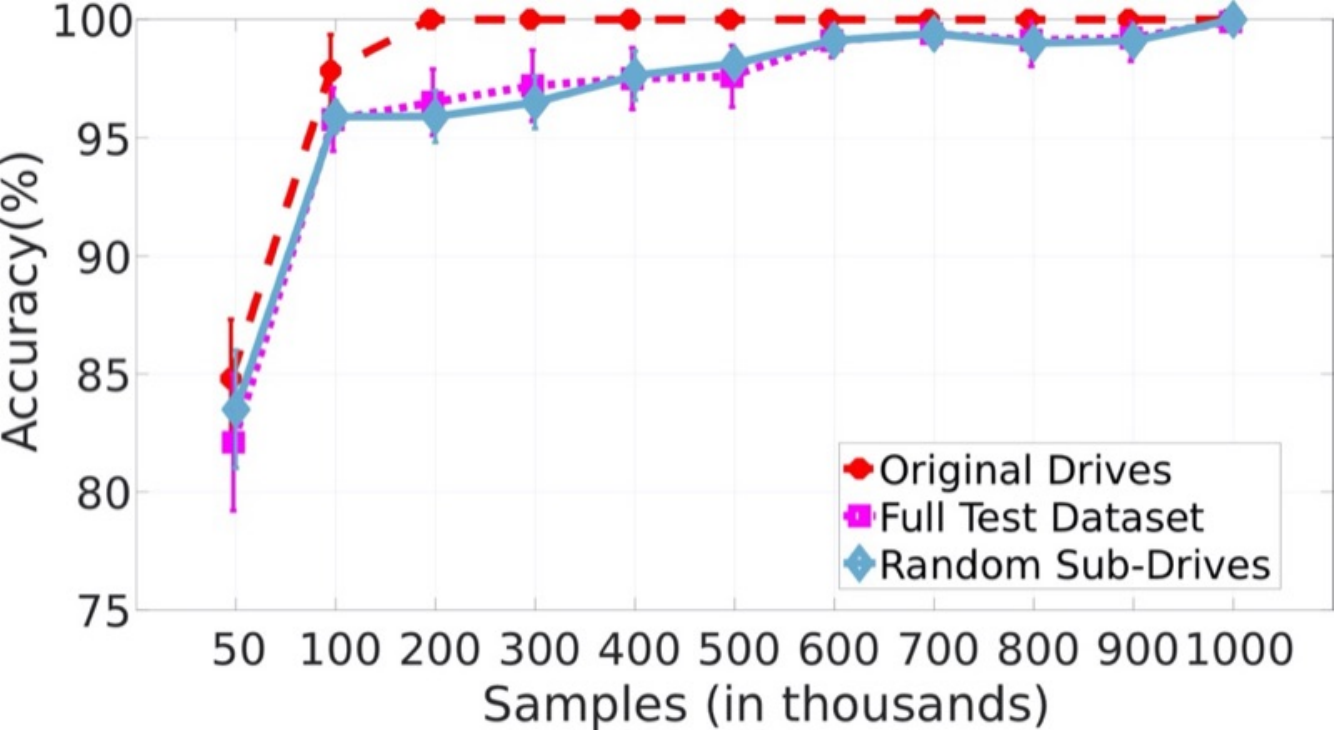}
    \label{fig:sjAcc}
    }
    \subfigure[Impact of PCI on Accuracy]{
    \includegraphics[width=0.31\textwidth]{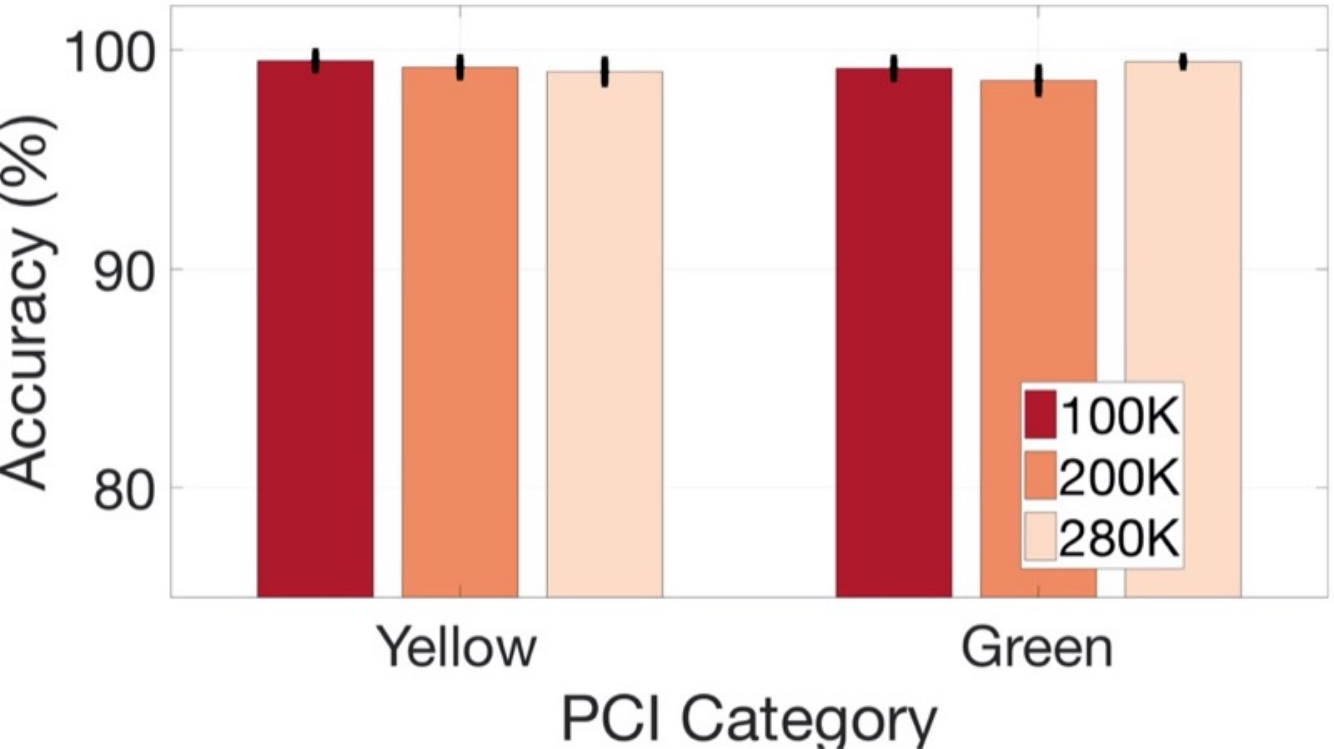}
    \label{fig:pci_acc}
    }
    \caption{\altname achieves 100\% lane classification accuracy in both SF (a) and SJ (b) on the original drives. Performance of the augmented dataset representing different speeds and vehicle variations trails the original test drives' performance closely. Accuracy increases over driving distance, with $90-95\%$ mean accuracy achieved with just $\sim100$~m of driving. We extract \textit{green} and \textit{orange} PCI sections of the route from San Jose (c) and show that (d) similar results are seen even with smoother roads.}%
\end{figure*}

We study \altname's ability to classify a vehicle's lane correctly as a function of the distance the vehicle has traveled in that lane. In both the SF and SJ routes, the model must distinguish between two adjacent lanes to do this. Figures~\ref{fig:sfAcc} and~\ref{fig:sjAcc} show the resulting model accuracy as a function of the number of accelerometer samples collected (which represents the driving distance in the lane). The average number of samples in the SF and SJ drives are approximately $1.2$M and $1.8$M; hence $100$K samples represent between $95$ and $130$~m  of driving distance. The maximum sample lengths we test against is the sample length of the quickest/shortest drive collected, which was $800$K for SF and $1$M for SJ.
As shown, \altname distinguishes between adjacent lanes with $100\%$ accuracy on the original (non-augmented) test drives within $200-300$K samples, and achieves over $90\%$ accuracy with just $100$K samples in both SF and SJ (corresponding to about $110$~m of driving distance). On both the original and the full, augmented test set, classification accuracy (averaged across drives for the original test set and across batches otherwise) increases with samples as expected, since longer driving distance on a lane yields more surface information to use in the differentiation. Note that the synthesized drives represent speed and various vehicle-specific variations not captured in the original drives; yet \altname classifies reliably on these as well, closely trailing the accuracy trend observed on the original dataset. We further introduce \textit{random sub-drive testing}, wherein we randomly pick $\ell$-sized sub-drives from each drive in the full test dataset. As shown, \altname continues to perform equally well on this, indicating its practicality for use in real-time driving wherein cars may provide the last $\ell$ samples from any point on the defined lane. For the rest of this section, \altname is evaluated using the random sub-drive method, thereby yielding results on the ``hardest'' testing scenario of the three discussed.

\subsection{Measuring Impact of Pavement Condition Indices}
The routes we drive in contain sections of different PCI categories including red, yellow and green that indicate poor, moderate and fair road conditions, respectively. This information is obtained by PCI scores provided in city websites~\cite{sf_pci,sj_pci} that are computed by surveying pavement distresses such as cracks, bumps and potholes. By capturing the range of PCIs in our dataset, we test \altname's performance in classifying between lanes of rough roads (in poor condition, presumably leading to more distinguished lane signatures) as well as smoother ones (presumably harder to distinguish). To assess if there is any significant decline in performance on smoother roads, we extract the green and yellow PCI subsections of the route (see Figure~\ref{fig:pci_png_sj}) from the SJ dataset. %
 Presumably, these roads of relatively better conditions may lead to less distinguishable lane signatures over short driving distances
 , which we now assess. %
 We train \altname with short sample lengths $\ell$ upto $280$K (the largest common sample length between the two PCI categories extracted from the drives). In Figure~\ref{fig:pci_acc}, we observe classification accuracy over $95\%$
for both these PCIs. In fact, \altname performs equally well even for lower sample lengths in each category, lending it suitable for roads of various conditions.

\subsection{Increasing the Number of Lanes} %

\begin{figure*}[htp]
\begin{minipage}[b]{\linewidth}
    \centering
    \subfigure[(Enlarged) ROC for Combines PCI Lanes]{
    \includegraphics[width=0.31\textwidth]{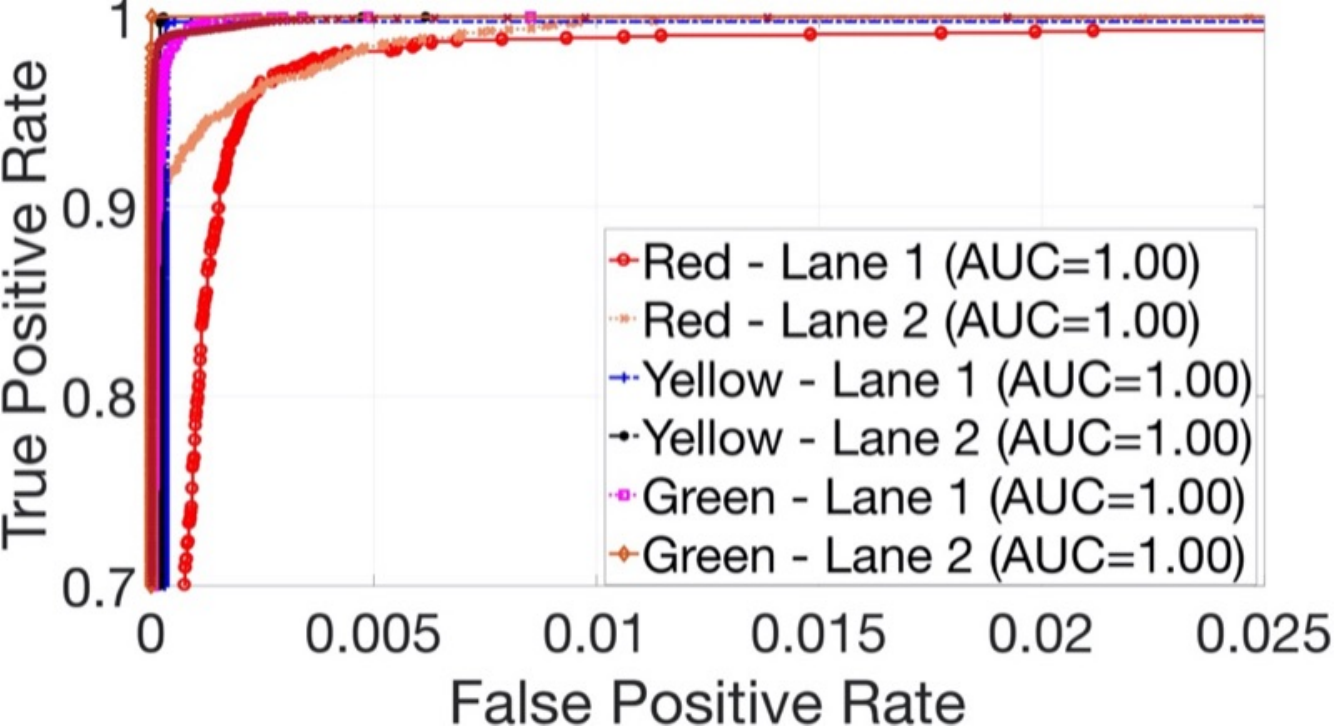}
    \label{fig:roc_pci}
    }
    \subfigure[(Enlarged) ROC for Combines SJ/SF Lanes]{
    \includegraphics[width=0.31\textwidth]{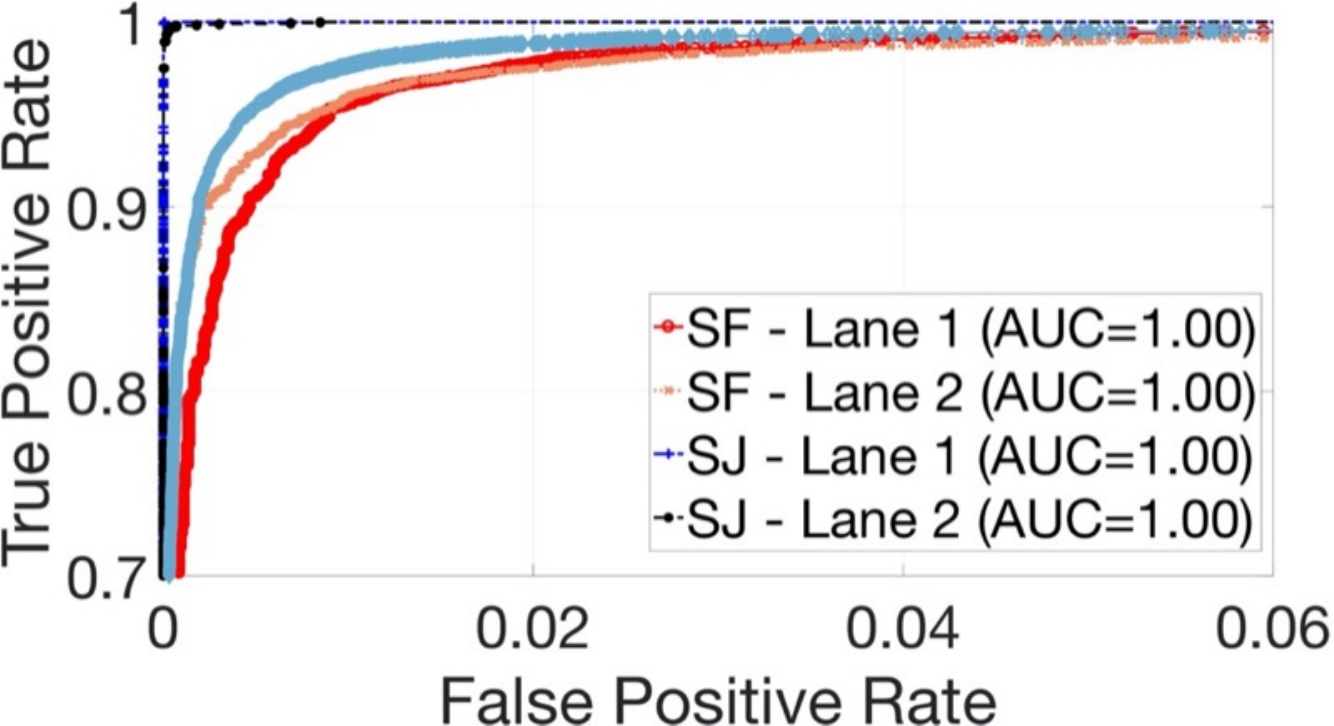}
    \label{fig:roc_sjsf}
    }
    \subfigure[F1 Score for Multi-Lane Combined Classification]{
    \includegraphics[width=0.31\textwidth]{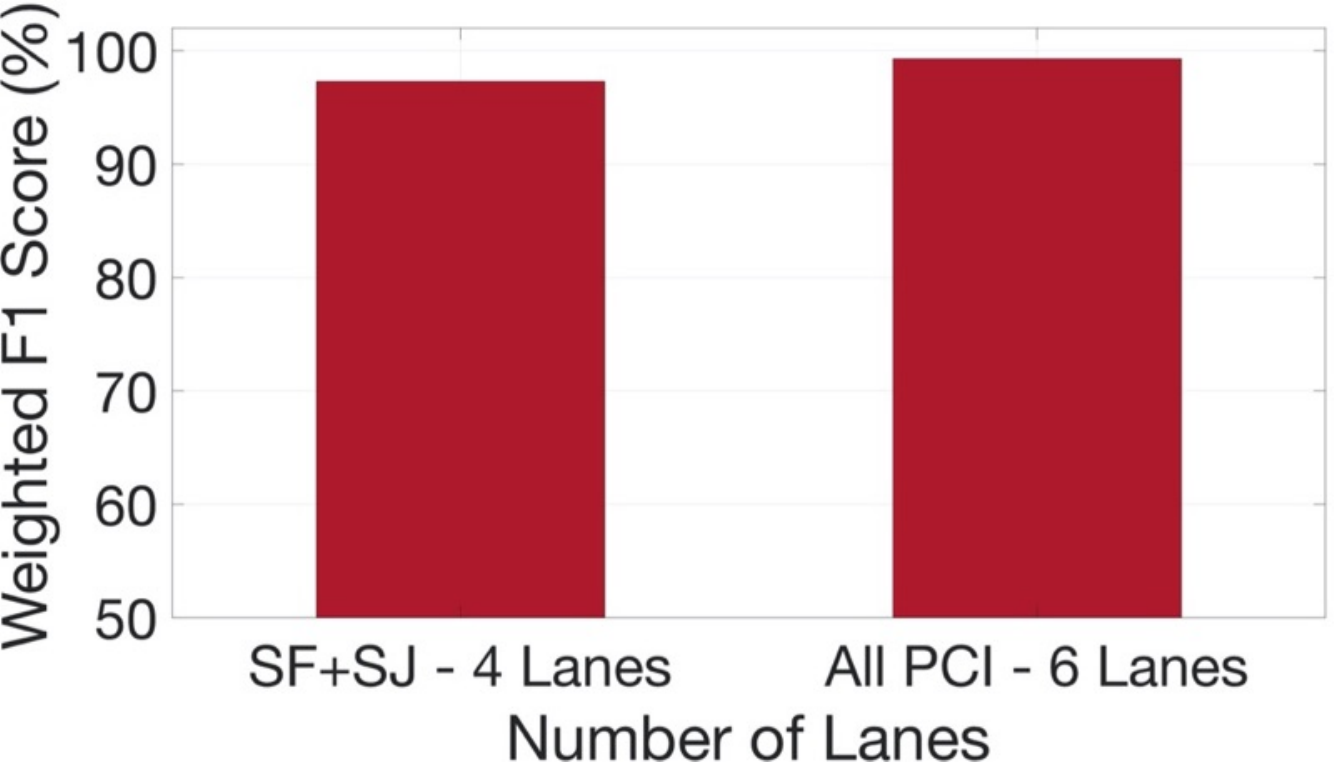}
    \label{fig:acc_alllanes}
    }
    \caption{Steep ROC curves with 100\% Area Under Curve for both PCI-based (a) and city-based (b) combined lane classification indicates that the model capacity is sufficient  to distinguish between multiple adjacent streets and for use in wide multi-lane highways. (c) The weighted F1 score in both cases classification exceeds $95\%$.}
\end{minipage}
\end{figure*}

While we've evaluated \altname on the binary task of distinguishing between two adjacent lanes, we now assess its performance on a larger number of lanes. High GPS errors~\cite{urbancanyon} often yield multiple candidate streets that the vehicle may be driving on. Figure~\ref{fig:Aerial_SF} shows such an example wherein GPS readings for our SF drives often map to adjacent streets. In such cases, \altname must distinguish between multiple lanes corresponding to all the candidate roads to accurately identify the vehicle's location. To assess whether our model has sufficient capacity to memorize and distinguish between a larger number of such lane signatures which are from similar/adjacent roads, we combine each PCI category's data from the SJ route to get a 6-lane dataset (2 of Green, Yellow and Red each). Figure~\ref{fig:roc_pci} shows the (zoomed-in) Receiver Operating Characteristics (ROC) observed for the resulting 6-lane classification task, with $\ell$ set to $280$K. The ROC illustrates the diagnostic strength of each class (corresponding to a lane) against all others as the discrimination threshold is varied. In this case, the ROC curves increase steeply, depicting true positive rate close to $1$ with false positive less than $0.01$, and almost $100\%$ area under the curve, i.e., the model almost always distinguishes between the positive and negative classes. To further assess if it can memorize much longer lane signatures for more than two lanes,e.g. as needed for use in large multi-lane highways, we generate a long 4-lane dataset by combining SF and SJ data (2 lanes of each route and over $1-2$~km each) %
and observe similar ROCs in Figure~\ref{fig:roc_sjsf}. The average F1 scores for both the 4-lane and 6-lane datasets is over $95\%$, as shown in Figure~\ref{fig:acc_alllanes}, weighted by the number of true instances for each label, thereby accounting for any label imbalance. %

\subsection{Measuring Performance During Lane Changes} 
We now shift our attention to more challenging lane-change scenarios where we test \altname's performance on drives with an artificially high frequency of lane changes (representing ``worst-case"). While we've ascertained that \altname distinguishes between adjacent lanes over both routes even within just $100$~m of driving distance, the model has insofar not encountered drives with lane change events wherein both lanes are present. This is more challenging to handle since \altname then models not only each lane's road surface signature individually, but also the \textit{adjacency} of the two lanes to learn, for instance, that a transition may occur from the $50^{th}$~m of Lane 1 to the $51^{st}$~m of Lane 2 but not to the $100^{th}$~m of Lane 2 since vehicles cannot (yet) teleport. 

\textbf{Generating Drives with Lane Changes.} To study this scenario, we start with the original drives that span one of two lanes in each route and construct new drives from these where lanes are alternated every $\alpha$ samples starting with either Lane 1 or Lane 2, for $\alpha \in \{25\text{K}, 50\text{K}, 100\text{K}, 200\text{K}, 300\text{K}, 400\text{K}, 500\text{K}, 600\text{K}, 700\text{K}\}$. These approximately represent the vehicle switching every $25$~m, $50$~m, $100$~m and so on. For SJ, we also include drives switching at $800$K and $900$K intervals since the minimum length across SJ drives is $1$M samples. After constructing these drives from the original data, we perform data augmentation to increase the dataset size; due to the time-warping technique therein, the resulting data-set of these lane-changing drives are of different lengths with lane changes happening at different locations. For each route, we train the \altname models on a combined dataset of the the original drives, augmented drives, stitched original drives, stitched augmented drives. We set $\ell$ to $800$K and $1$M for SF and SJ respectively. We modify the proposed loss function in (\ref{eq:lossEq}) to weight each cell's output uniformly (rather than increasing their weights in accordance with their proximity to the final cell). Since the drives in this dataset contains lane changes, LSTM cells do not necessarily provide more information than preceding ones to classify any single lane that the drive may have started out in or switched to earlier.

\textbf{Ground Truth Labeling.} 
\begin{figure*}
    \centering
    \includegraphics[width=.8\textwidth]{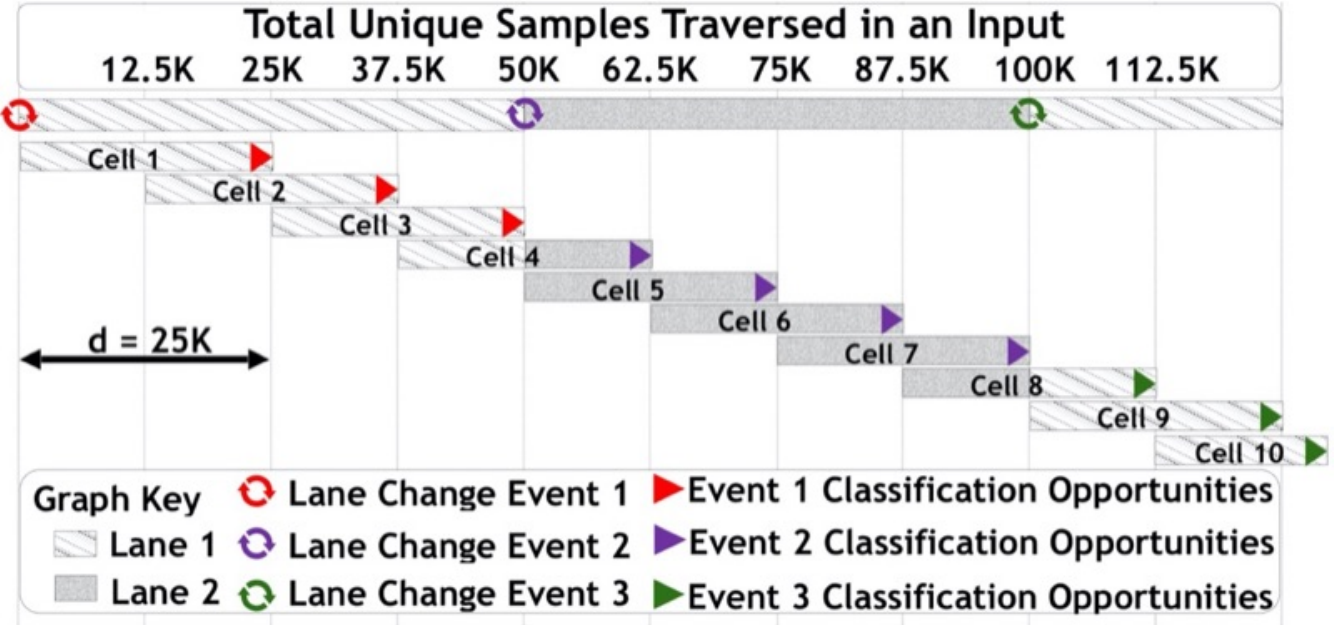}
    \caption{We show how to interpret individual cell outputs from the model's final LSTM layer when the input drives exhibit periodic lane changes. We consider a drive where the lane is alternated every $50$K samples (i.e. every $50-60$~m) and each LSTM cell corresponds to $25$K input samples with $12.5$K stride (or $12.5$K overlapping samples between adjacent cells). As seen, after each such lane change, there are three to four opportunities for classifying the new lane correctly, corresponding to the outputs of cells that end with samples of the new lane. When the next lane change event happens, the classification opportunities for the previous one are over.}
    \label{fig:laneSwitchingIllus}
\end{figure*}
To understand how ground truth labeling is done for the lane-changing drives, we refer to the illustration in Figure~\ref{fig:laneSwitchingIllus}. \textit{Note that all subsequent mention of "cells" refers to the LSTM cells in the output layer of the model.} Consider a constructed drive where the lane is alternated every $50$K samples. Consider $d=25$K, i.e., each LSTM cell output layer corresponds to $25$K input samples (see Figure~\ref{fig:lstm_arch}). If the drive starts out in Lane $1$ and makes a lane change at $50000^{th}$ sample, the first three cells would entirely span this Lane 1 portion, with Cell 1 corresponding to samples $0-25000$, Cell 2 corresponding to samples $12500-37500$ (due to a $12.5$K stride), and Cell 3 corresponding to samples $25000-50000$. Since the lane is switched to Lane 2 at $50000^{th}$ sample, Cell 4 then contains $25000$ samples where the first $12500$ samples correspond to the last $12500$ samples from Lane 1 and the next $12500$ samples correspond to the first $12500$ from the switched point of Lane 2. This is illustrated in Figure~\ref{fig:laneSwitchingIllus}. 

To study the lane classification accuracy in the presence of these lane change events, we inspect the intermediate LSTM cells of the output layer (not just the final one) to assess, for each lane change event, whether the new lane was correctly classified before the lane was changed again. The final cell's output only provides information on the last lane segment that was switched to and no visibility into classification performance on the intermediate points in the road where the lane was switched. We hence need to determine per-cell ground truth to assess the classification accuracy at the output of each cell.

There are two ways to compute per-cell ground truth. First, we may choose the lane that has the larger representation in the input samples as ground truth, which we refer to as the \textit{Most Frequent} (MF) labeling policy. For instance, if a cell corresponds to a $25$K sample where the first $20$K samples belong to Lane 1 and the latter $5$K to Lane 2, it may be appropriate to mark the ground truth for this cell at Lane 1 to aid the model during training time in understanding this input as primarily a Lane 1 pattern. If both lanes form exact halves of the $25$K sample, as in Cell 4 in Figure~\ref{fig:laneSwitchingIllus} for example, the more recent lane is chosen as the ground truth (i.e. Lane 2 in the case of Cell 4). On the other hand, from the vehicle's perspective, as soon as it switches lanes, any classification result indicating the previous lane is incorrect, even if the vehicle has only traversed $5000$ samples in the new lane and still retains $20000$ samples from the previous one. Hence, the second mechanism may be to mark the ground truth for such a sample as Lane 2, which we refer to as the \textit{Last Occurrence} (LO) labeling policy. However, it is less clear then what the model learns about lanes in this case if a sample containing $80\%$ of Lane 1 surface patterns is marked as Lane 2. 

We navigate this design decision by studying the impact of both these methodologies. Note that LO is the only policy used during test time - i.e. if a vehicle's lane is ever classified as its previous one, this is incorrect regardless of the proportion of leftover samples from the previous lane in the model input. The design decision is to choose the best labeling policy during train time that would result in the best LO-based classification during test time.

\begin{figure*}[!t]
    \centering
    \subfigure[SF - Accuracy over varying frequencies of lane changes]{
    \includegraphics[width=0.47\textwidth]{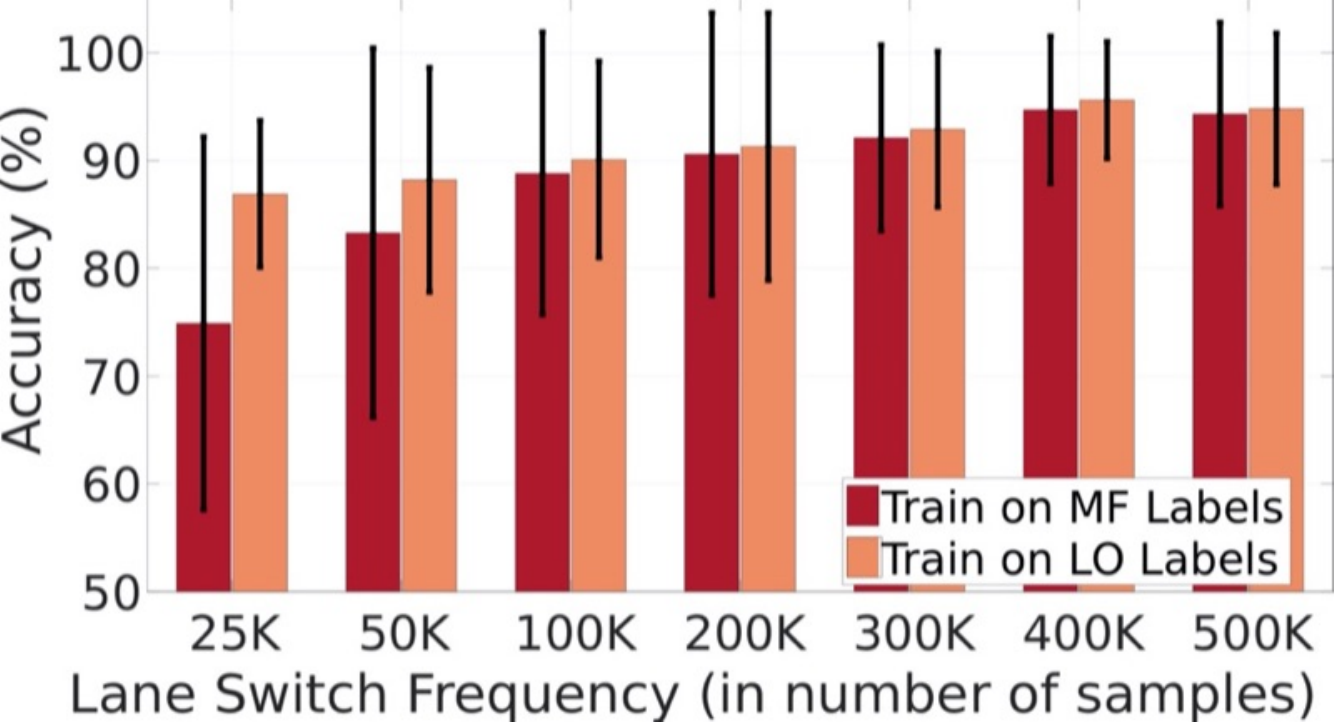}
    \label{fig:SFLaneSwitch}
    }
    \subfigure[SJ - Accuracy over varying frequencies of lane changes]{
    \includegraphics[width=0.47\textwidth]{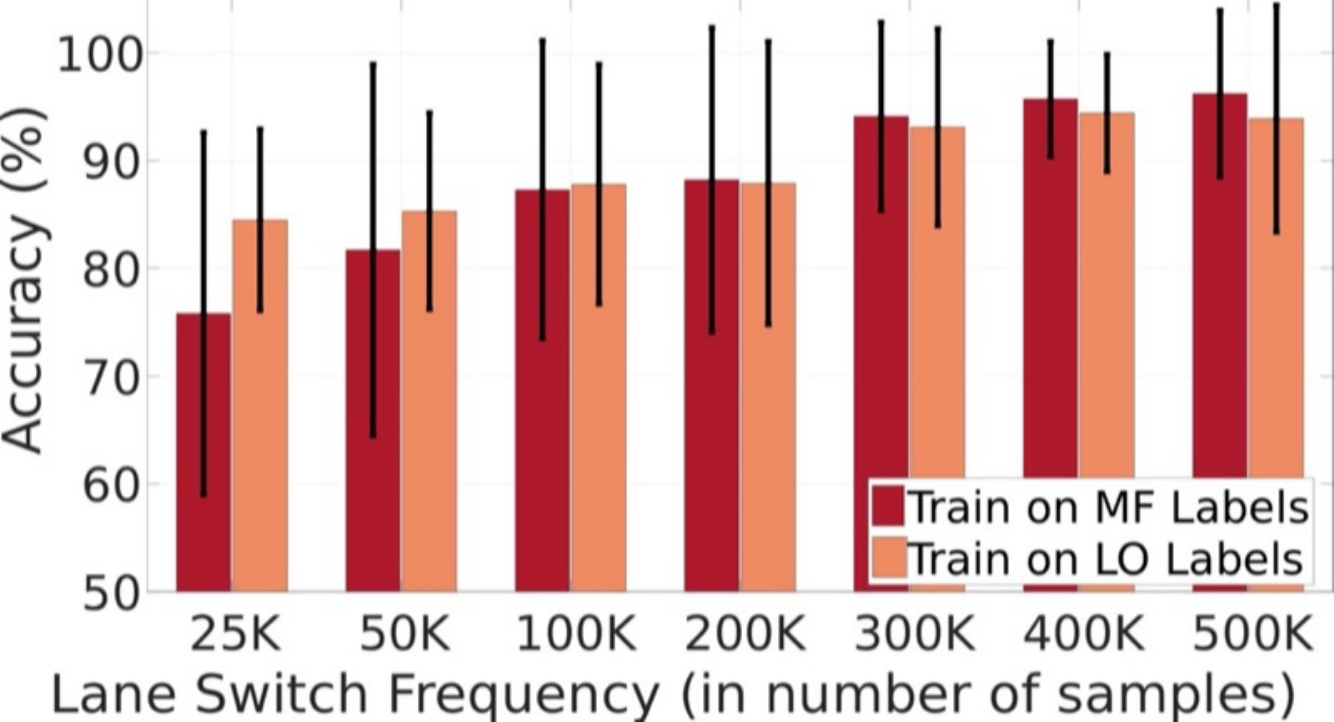}
    \label{fig:SJLaneSwitch}
    }
    \caption{Across both routes, using the Last Occurrence labeling policy for ground truth during training time yields higher mean classification accuracy and lower variance for drives with very frequent lane changes. However, the Most Frequent policy results in equivalent or even better performance in the case of the SJ route for less frequent lane change events. }
    \label{fig:laneSwitchEvents}
\end{figure*}

Figure~\ref{fig:laneSwitchEvents} shows the accuracy of lane detection for drives with lane switch frequencies of $25-300$K samples, averaged across batches and cells of the output layer. Note that the choice of $d$ in the model is crucial for handling these lane-change scenarios. Since drives may switch lanes as frequently as every $25$K samples, we set $d=25$K, i.e., each LSTM cell corresponds to $25$K input samples (see Figure~\ref{fig:lstm_arch}) so that atleast one cell is guaranteed to make a prediction for data from the switched lane before another lane change occurs. As seen in Figure~\ref{fig:laneSwitchEvents}, the LO policy results in almost $90$\% accuracy for drives that switch lanes even as frequently as every $\sim25$ meters for both the SF and SJ routes, while the MF policy achieves much lesser. However, note that accuracy with MF labeling increases steadily as the lane switch frequency decreases. %
 However, since seemingly outperforms MF for both routes, there may be no reason to consider the MF labeling technique in practice; however, note that in SJ, LO starts to perform worse as lane changes occur less frequently, i.e., as vehicles spend longer distances in a switched lane. Since the MF technique associates predominant lane patterns in the input with the actual lane they belong to, it directly yields more accurate results the longer the vehicle is on a lane after a lane change event. The LO technique, though achieving more accurate lane identification with $25$~m after a lane change, begins to have poor performance over longer distances in SJ as the model does not seem to have learnt a clear differentiation between Lane 1 and Lane 2 patterns in this route. We now inspect this further.

\textbf{Timeliness of Classification.} While Figure~\ref{fig:laneSwitchEvents} shows the average classification accuracy across for drives (of $800$K/$1$M samples) with constant lane switches, it does not indicate \textit{the timeliness} of model response to lane change events. In other words, we now analyze \textit{how long}, as measured by number of samples, after a lane change event the model classification is updated to indicate the new lane. To do this, we define \textit{Windows of Classification Opportunities}, which refers to certain points after a lane change event where an opportunity arises to identify that there was a lane switch and correctly classify the new lane. For instance, in the example discussed above from Figure~\ref{fig:laneSwitchingIllus}, there are \textit{three classification opportunities} for the vehicle driving in Lane 1 corresponding to the output of Cells 1, 2 and 3, before the vehicle changes lane and the classification opportunity for the previous road segment is lost. Note that even though half of Cell 4 spans the previous lane, since a lane change occurs \textit{within Cell 4} (i.e. in the input samples that correspond to Cell 4), Cell 4 does not provide a classification opportunity for the previous segment. Even if the \altname model predicts Cell 4 to be Lane 1, that is considered an incorrect prediction since, from the perspective of the driver using \altname, the vehicle has switched lanes and hence the previous lane is now outdated (a.k.a LO labeling). Similarly, for the distance spanning $50000-100000^{th}$ sample, when the vehicle is in Lane 2, there are 4 classification opportunities as shown, before the lane change event happens at $100$K and the classification opportunity for the new lane segment starts at Cell 8 containing $12.5$K samples of the changed lane. We further define the \textit{distance of a classification window} to be the length (in samples) after the lane change event occurs when the window opens up. For instance, for the drive spanning samples $1-50000$ in Lane 1 in Figure~\ref{fig:laneSwitchingIllus}, the first classification window at Cell 1 arises at $25000$ samples from the event, the second window arises at $37500$ samples from the event, and the third window arises at $50000$ samples from the event. For the second segment of the drive spanning samples $50000-100000$, the first window for classification opportunity at Cell 4 arises at $12500$ samples after the event happened, second window arises at Cell 5 at $25000$ samples after the event, third at Cell 6 at $37500$ samples from the event, and fourth at Cell 7 at a distance of $50000$ samples from the event.

\begin{figure*}[!t]
    \centering
    \subfigure[SF - Performance of Model Trained with MF Labels on Classification Windows for $50$K Lane Change Frequency]{
    \includegraphics[width=0.31\textwidth]{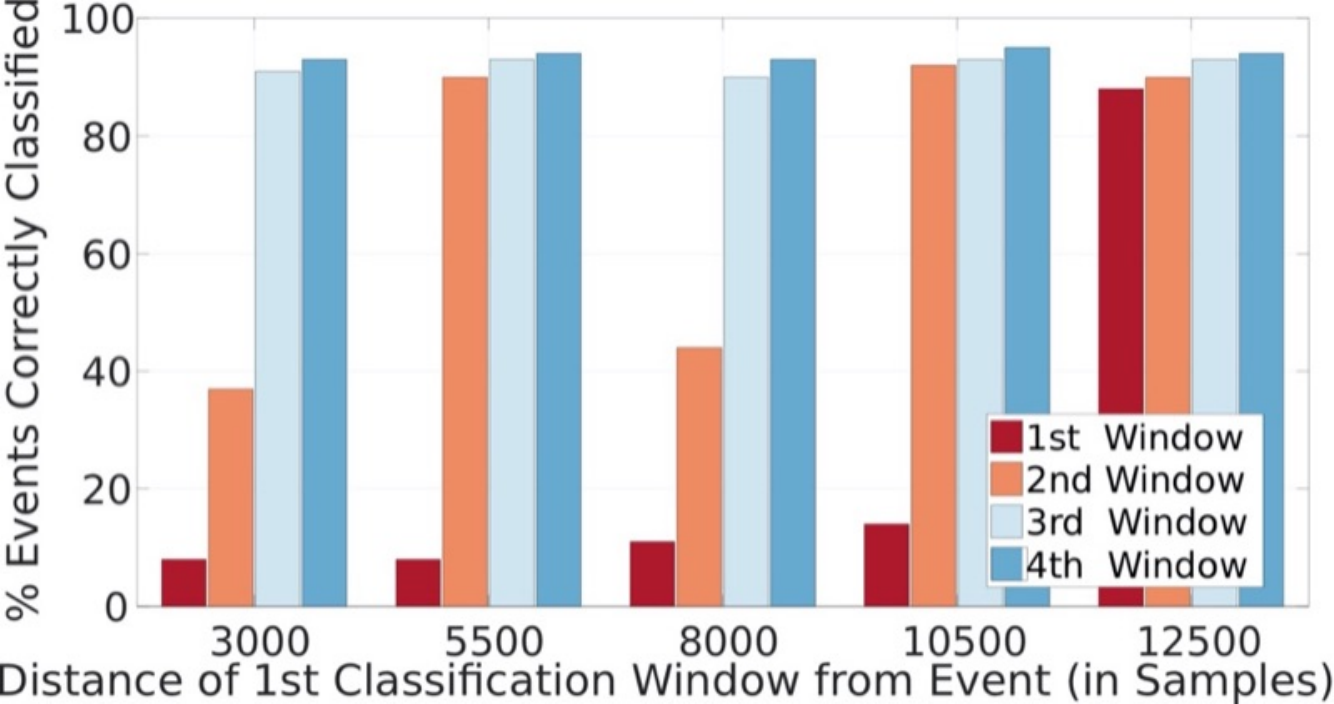}
    \label{fig:SFUtilizationOfWindowsAllMF}
    }
    \hfill
    \subfigure[SF - Performance of Model Trained with LO Labels on Classification Windows for $50$K Lane Change Frequency]{
    \includegraphics[width=0.31\textwidth]{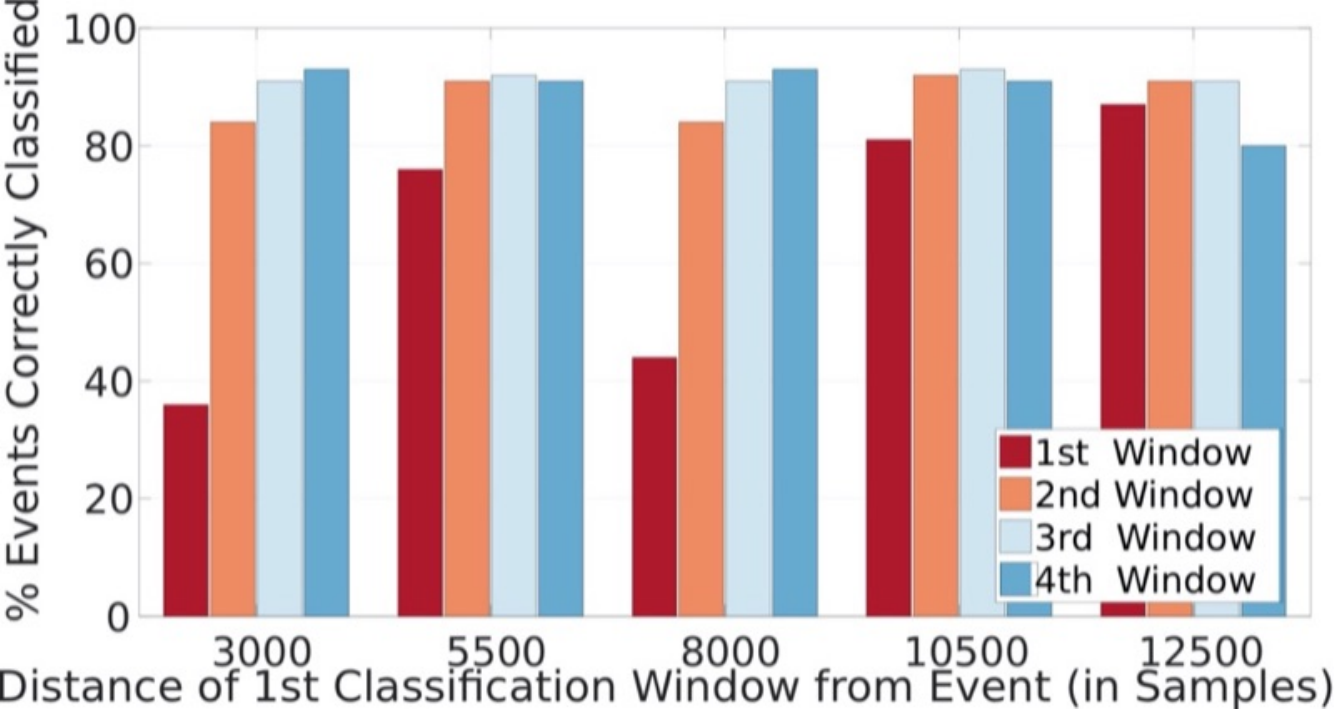}
    \label{fig:SFUtilizationOfWindowsAllLO}
    }
    \hfill
    \subfigure[SF - Model Performance on the first ten Classification Windows averaged across $25-500$K Lane Change Frequencies]{
    \includegraphics[width=0.31\textwidth]{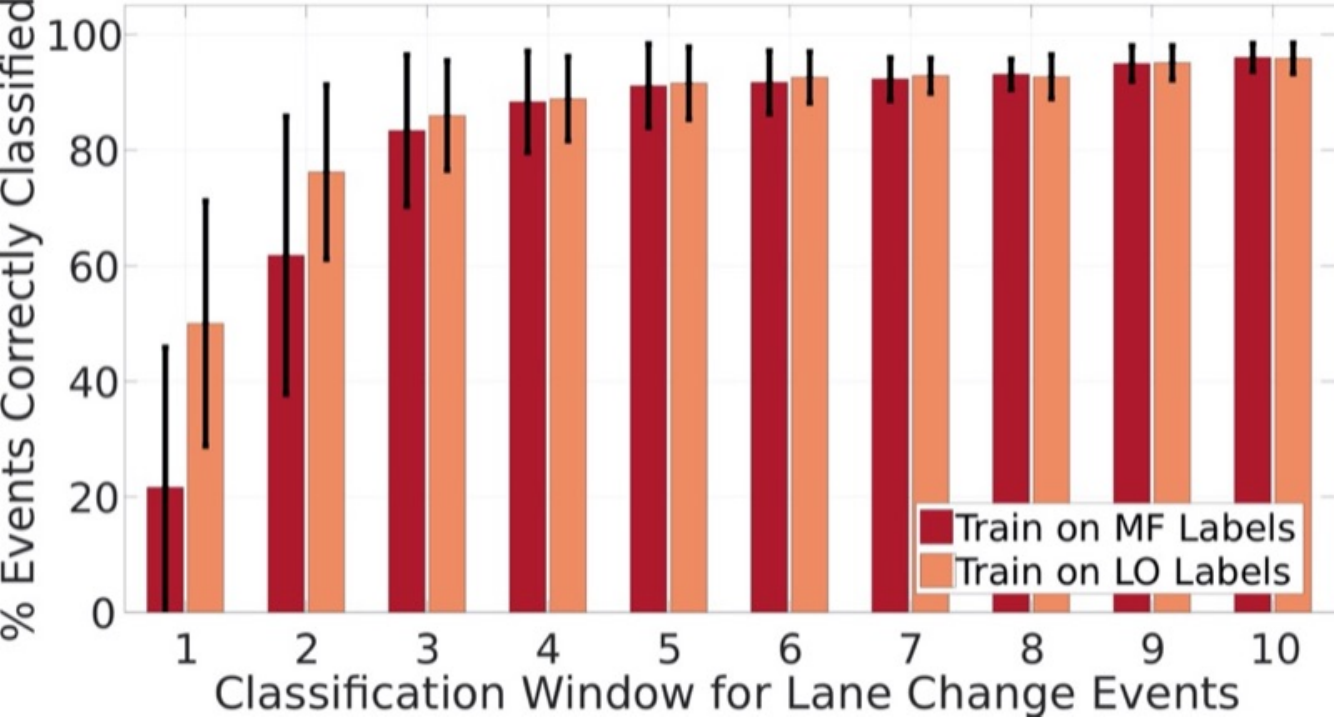}
    \label{fig:SFAllStitchWindow}
    }
    \caption{(a) Training with MF labels typically achieves poor classification accuracy in the first classification window following an event, though the subsequent ones steadily increase in accuracy. After two classification windows, the accuracy is above 90\%. (b) In comparison, training with LO labels results in much higher classification as quickly as the first classfication window following a lane change event, even when the window arises just 3000-5000 samples after it (<5 meters). However, accuracy seems to drop for 3rd and 4th windows. (c) When averaged across drives of all lane change frequencies, LO outperforms MF for earlier windows and performs equally well for later ones.}
    \label{fig:laneSwitchEventsSF}
\end{figure*}

\begin{figure*}[!t]
    \centering
    \subfigure[SJ - Performance of Model Trained with MF Labels on Classification Windows for $50$K Lane Change Frequency]{
    \includegraphics[width=0.31\textwidth]{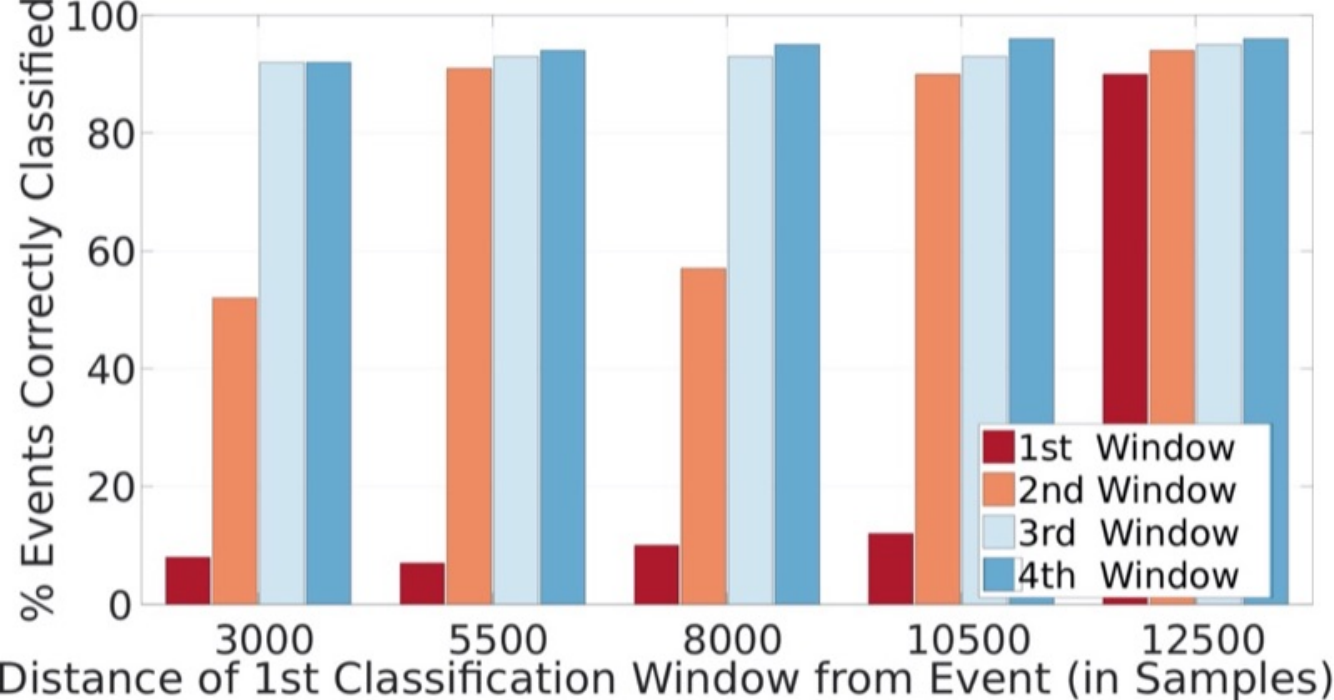}
    \label{fig:SJUtilizationOfWindowsAllMF}
    }
    \hfill
    \subfigure[SJ - Performance of Model Trained with LO Labels on Classification Windows for $50$K Lane Change Frequency]{
    \includegraphics[width=0.31\textwidth]{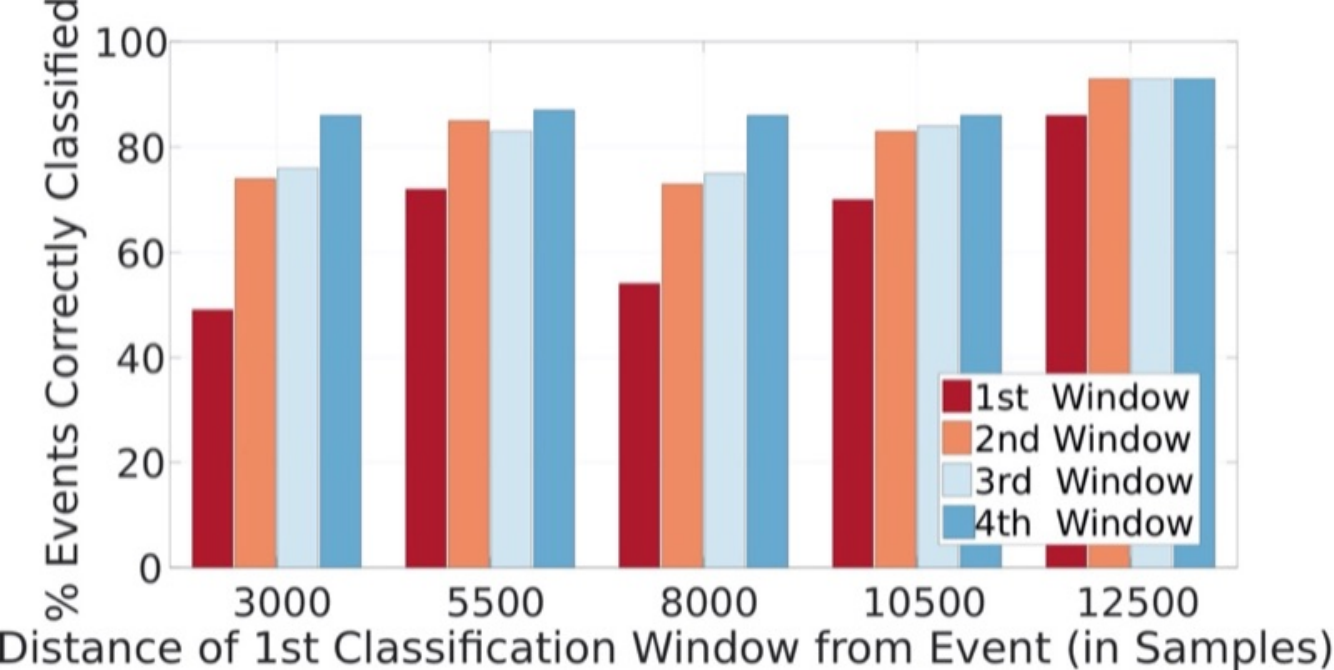}
    \label{fig:SJUtilizationOfWindowsAllLO}
    }
    \hfill
    \subfigure[SJ - Model Performance on the first ten Classification Windows averaged across $25-500$K Lane Change Frequencies]{
    \includegraphics[width=0.31\textwidth]{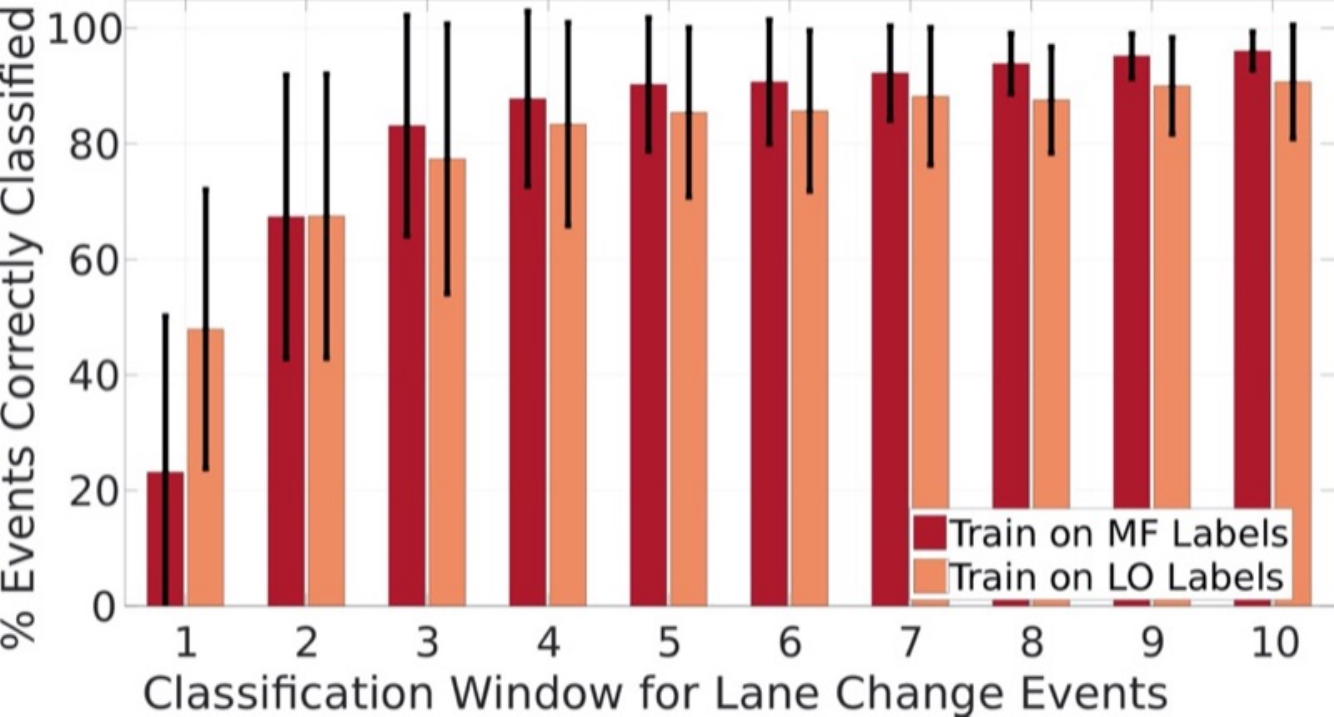}
    \label{fig:SJAllStitchWindow}
    }
    \caption{We show the tradeoff between long-term classification accuracy and classification timeliness that emerges in the SJ model. As seen  (b,c), training with MF labels achieves higher classification over increasing classification windows from the event, while training on LO labels yields quicker lane change detection (i.e. high classification accuracy even on the first classification window following the event) but poorer performance over distances. Within $\sim25$~m of driving after lane change, between $98-65$\% lane detection accuracy is achieved. (c) When performance is averaged across drives of all stitches, it is evident that LO yields poorer accuracy on longer distances within a lane.}
    \label{fig:laneSwitchEventsSJ}
\end{figure*}

To study the timeliness of lane change detection, i.e. how quickly after a lane change event the model correctly identifies the new lane, we analyze the model performance on each classification window of a lane change event. For instance, for drives that switch lane every $50$K samples, each lane change event has $4$ classification windows (also evident from Figure~\ref{fig:laneSwitchingIllus}) that occur at $12.5$K, $25$K, $37.5$K and $50$K samples after the event (except for the first event). %
In practice, however, vehicles may switch lanes at any point in the road and not at specific pre-determined 50K switch points. The augmented data for the $50$K drives capture this scenario as the time-warping process stretches/shrinks the samples, hence placing the lane change location at different points of the road. When testing on this entire $50$K dataset, we hence see classification windows occur at varied intervals. While previously, the first classification window always occurred at $12.5$K samples from the event, now we see the first classification window for different lane change events in the drive occurring at $3000$, $5000$, $8000$, $10500$ and $12500$ samples from the event. Subsequent classification opportunities after the first window always occur at incremental $12.5$K samples since cell stride, or $d$, is $12.5$K. Note that not all lane change events now have $4$ classification windows: due to time-warping, some lane segments get shrunk and therefore switch even sooner than $50$K. 

We group lane change events based on when their first classification window occurs and study the utilization of classification opportunities within each of these groups. In Figures~\ref{fig:SFUtilizationOfWindowsAllMF} and ~\ref{fig:SJUtilizationOfWindowsAllMF}, the model has been trained on the SF and SJ route respectively with MF labeling technique. We see that irrespective of when the first classification window occurs, there is a steady increase in accuracy across classification windows after an event, in keeping with the general observation in Figure~\ref{fig:laneSwitchEvents}. Also irrespective of when the first classification window occurs, the first one is consistently poorly utilized unless the first window occurs at $12.5$K samples, indicating that atleast $50\%$ of the input sample for the cell must be in the new lane for the model to classify it correctly (which is in line with the MF labeling it was trained on). Figures~\ref{fig:SFUtilizationOfWindowsAllMF} and~\ref{fig:SJUtilizationOfWindowsAllMF} also show that the second classification window is better utilized when larger number of samples are provided for the first despite the utilization of the first typically being poor, indicating that the internal cell context more accurately reflect the newer lane with larger samples in the cell even if the cell output itself incorrect. Depending on the distance of the first classification window from the lane-change event, $40-80$\% classification accuracy is reached by the second classification window, i.e., between $\sim15-50$ meters of driving in the new lane. All third and fourth classification windows (i.e. after $\sim40-50$~m of lane change) have consistently high accuracy. %

Figures~\ref{fig:SFUtilizationOfWindowsAllLO} and ~\ref{fig:SJUtilizationOfWindowsAllLO} show the performance of the SF and SJ models trained with LO labels. With LO labels, about $40$\% classification accuracy is achieved just in the first classification window for almost all distances of the first window, and $60-90$\% for the second window (i.e. within $\sim15-50$ meters of driving in the new lane. Since the model has been trained on the same type of labeling process that it is tested on, this significant improvement is performance is reasonable. However, in this case, \textit{the utilization of the fourth window decreases} across certain first window distances for SJ. This drop in accuracy across subsequent classification windows is especially pronounced when the first window distance is $12.5$K samples, indicating that the latter half of the lane segment is more likely to be incorrectly classified. This is a casualty of training with the LO labels, also seen in Figure~\ref{fig:SJLaneSwitch}; since this section of the lane may have been marked incorrectly when clubbed with the other lane in some training samples (due to the LO process), \altname's modeling of lane-specific inherent surface features is sometimes compromised. Figures~\ref{fig:SFAllStitchWindow} and~\ref{fig:SJAllStitchWindow} show the model performance on the first ten classification windows for lane change events (if the windows exist), averaged across drives of 25-500K switch frequency. As we see for SJ, there is clear performance degradation with the LO technique over the longer driving distances (i.e. farther windows of opportunity). \textit{In other words, with LO vs MF, we trade off between timeliness of lane change detection and long-term accuracy of classification.} As seen, the trade-off may not be equally pronounced for all roads; in our particular case, the SJ route exhibits this more.

\begin{figure}[!t]
    \centering{
    \includegraphics[width=0.47\textwidth]{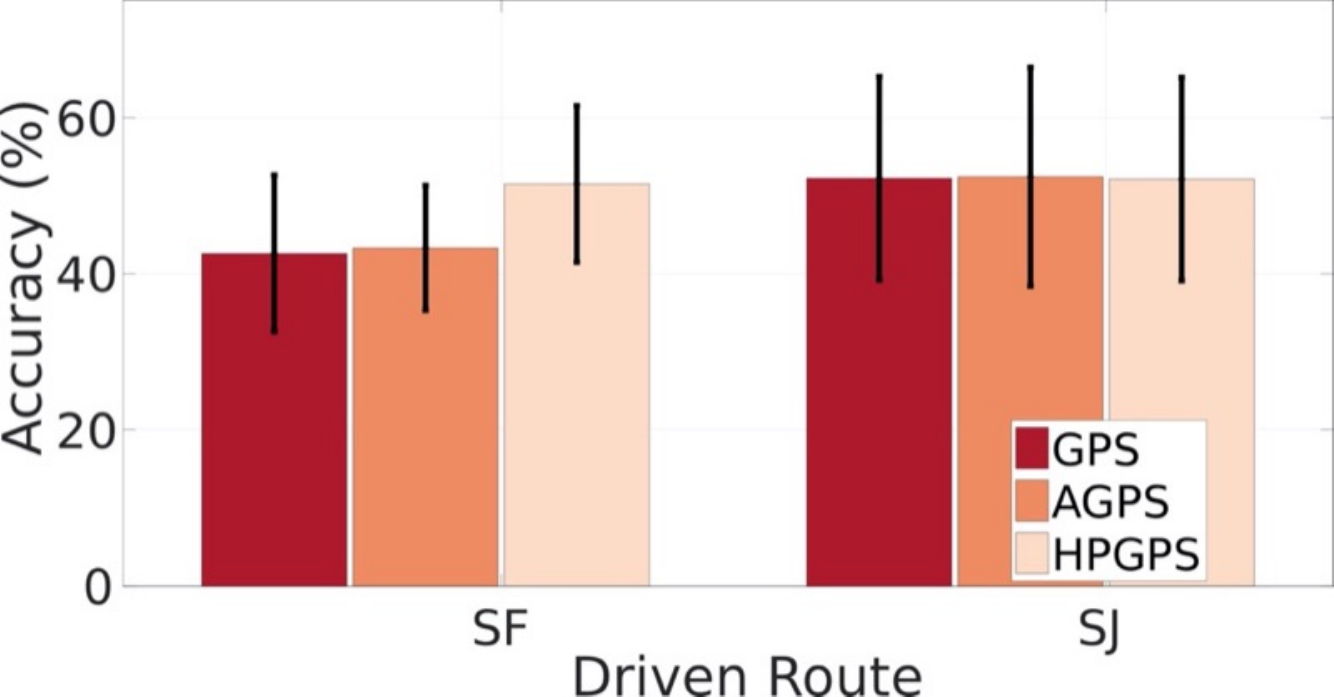}
    }
    \caption{We see that the baseline GPS traces yield only about 50\% lane classification accuracy, even in the case of HPGPS, which \altname clearly outperforms.}
    \label{fig:baselineAcc}
\end{figure}

\textbf{Baseline Performance Amidst Frequent Lane Changes.} Finally, we obtain the lane-detection accuracy yielded by the baseline sensors (GPS/AGPS/HPGPS) to compare with \altname. To do this, we use OpenStreetMap (OSM)~\cite{osm} to perform our final \textit{Lane-level Map Matching} with the baseline traces. Since current OSM data only includes a single set of \textit{nodes} per road segment, we populate lane-level \textit{nodes} per road using Java OpenStreetMap Editor (JOSM)~\cite{josm}, at $10$~m interval nodes per lane for both routes. Figure~\ref{fig:baselineAcc} shows the results; as expected, traditional map-matching on these coarse readings does not yield waypoints of sufficient granularity to pinpoint vehicles' lanes. 

Not even HPGPS yields mean accuracy over about 50\% on either routes, indicating that no better than a random guess can be made for lane identification with the GPS data. We note that even though the traces depicted in Figures~\ref{fig:Aerial_SF} and \ref{fig:Aerial_SJ} look less noisy in San Jose than San Francisco, the GPS error is nevertheless prohibitively high to achieve lane-level localization.

\subsection{Excavating the Model for Road Surface Insights}
We now show that \altname can be used to provide an extremely useful view into the road surface structure by essentially characterizing the achievable classification accuracy along different sections of the road at different distance by training just one model. To illustrate this, we consider the dataset of drives without lane changes (original as well as augmented) and hence also use the weighted loss function in (\ref{eq:lossEq}).

\setlength{\tabcolsep}{2pt}
\begin{table*}[!ht]
    \centering
\begin{minipage}[t]{\textwidth}\centering
\caption{Testing Accuracy From Intermediate Cells - SF}
\label{tab:sf_lstm}
\vspace{-1.5em}
  \begin{threeparttable}
\begin{tabular}{|l|lllllllllllllll}
\textbf{100} & 82.6 \tiny{1.9} &  &  &  &  &  &  &  &  &  &  &  &  & &  \\ %
\textbf{200} & 82.0 \tiny{1.7} & 88.7 \tiny{1.4} &  92.3 \tiny{1.4} \\ %
\textbf{300} & 80.2 \tiny{2.1} & 87.9 \tiny{1.6} &  91.8 \tiny{1.4} & 94.2 \tiny{1.3} & 95.5 \tiny{1.2} \\ %
\textbf{400} & 77.1 \tiny{1.7} & 86 \tiny{1.2} &  90.3 \tiny{1.3} & 93.6 \tiny{1.1} & 95.7 \tiny{1.0} & 96.8 \tiny{.7} & 97.3 \tiny{.6} \\ %
\textbf{500} & 75.1 \tiny{1.9} & 84.1 \tiny{1.6} &  89.2 \tiny{1.4} & 93.1 \tiny{1.1} & 95.2 \tiny{.9} & 96.3 \tiny{.8} & 96.7 \tiny{.7} & 97.0 \tiny{.7}  & 97.2 \tiny{.6} \\ %
\textbf{600} & 74.6 \tiny{1.9} & 84.6 \tiny{1.7} &  90.1 \tiny{1.6} & 93.5 \tiny{1.2} & 95.7 \tiny{1.2} & 96.86 \tiny{1.2} & 96.8 \tiny{1.0} & 96.7 \tiny{1.0}  & 96.6 \tiny{.8} & 96.3 \tiny{.9} & 97.1 \tiny{.8} \\ %
\textbf{700} & 79.7 \tiny{2.06} & 86.2 \tiny{1.76} &  91.1 \tiny{1.6} & 94.7 \tiny{1.41} & 96.9 \tiny{1.19} & 97.9 \tiny{.9} & 98.4 \tiny{.69} &98.9 \tiny{.56}  & 99.2 \tiny{.38} & 99.3 \tiny{.38} &99.4 \tiny{.37}  & 99.4 \tiny{.41} & 99.4 \tiny{.36} \\
\textbf{800} & 78.7 \tiny{2.3} & 86.6 \tiny{2.1} 
&  91.4 \tiny{1.7} & 
95.2\tiny{1.4} & 
97.0 \tiny{.9} & 
98.0 \tiny{.7} & 
98.4 \tiny{.6} &
98.6 \tiny{.5}  & 
98.8 \tiny{.4} & 
98.9 \tiny{.4} &
99.1 \tiny{.4}  & 
99.2 \tiny{.4} & 
99.4 \tiny{.4} & 
99.3 \tiny{.4} & 99.3\tiny{.4} \\ \hline
& \textbf{100} & \textbf{150} & \textbf{200} & \textbf{250} & \textbf{300} & \textbf{350} & \textbf{400} & \textbf{450} & \textbf{500} & \textbf{550} & \textbf{600} & \textbf{650} & \textbf{700} & \textbf{750} & \textbf{800} \\ \hline
\end{tabular}
\begin{tablenotes}
      \item Note:  Trained sample lengths $\ell$ (first column) and accumulated sample length at each intermediate output (last row) are in thousands. Variance specified next to mean \footnotesize{in smaller font}.
\end{tablenotes}
\end{threeparttable}
\end{minipage}
\end{table*}

\setlength{\tabcolsep}{.5pt}
\begin{table*}[!ht]
    \centering
\begin{minipage}[t]{\textwidth}\centering
\caption{Testing Accuracy From Intermediate Cells - SJ}
\label{tab:sj_lstm}
\vspace{-1.5em}
  \begin{threeparttable}
\begin{tabular}{|l|lllllllllllllllllll}
\textbf{100} & 83.9 \tiny{2.3} &  &  &  &  &  &  &  &  &  &  &  &  & &  \\ %
\textbf{200} & 81.5 \tiny{1.9} & 88.9 \tiny{1.4} &  92.3 \tiny{1.1} \\ %
\textbf{300} & 78.8 \tiny{1.5} & 87.6 \tiny{1.9} &  93.0 \tiny{1.4} & 95.3 \tiny{1.1} & 96.5 \tiny{1.1} \\ %
\textbf{400} & 77.1 \tiny{1.8} & 85.3 \tiny{1.0} &  90.2 \tiny{1.0} & 93.2 \tiny{1.5} & 95.2 \tiny{1.2} & 96.8 \tiny{1.2} & 97.4 \tiny{1.2} \\ %
\textbf{500} & 75.7 \tiny{1.8} & 83.5 \tiny{1.4} &  89.7 \tiny{1.1} & 93.4 \tiny{1.5} & 95.6 \tiny{1.1} & 96.5 \tiny{1.0} & 97.3 \tiny{.6} & 97.6 \tiny{.7}  & 97.7 \tiny{.6} \\ %
\textbf{600} & 74.5 \tiny{3.1} & 83.2 \tiny{2.2} &  89.1 \tiny{1.6} & 93.2 \tiny{.6} & 96.2 \tiny{1.2} & 98.0 \tiny{.8} & 98.6 \tiny{.6} & 98.8 \tiny{.6}  & 98.9 \tiny{.5} & 99.0 \tiny{.5} & 99.2 \tiny{.5} \\ %
\textbf{700} & 
80.3 \tiny{2.0} & 
87.8 \tiny{1.7} & 92.3 \tiny{1.7} & 95.3 \tiny{1.0} & 97.1 \tiny{.9} & 97.9 \tiny{.7} & 98.5 \tiny{.6} &99.1 \tiny{.4}  & 99.4 \tiny{.4} & 99.6 \tiny{.2} &99.5 \tiny{.3}  & 99.6 \tiny{.3} & 99.4 \tiny{.3} \\
\textbf{800} 
& 78.1 \tiny{2.1} & 86.5 \tiny{2.2} &  90.9 \tiny{1.9} & 93.6 \tiny{1.7} & 95.4 \tiny{1.5} & 97.3 \tiny{1.0} & 98.3 \tiny{.7} & 99.1 \tiny{.4}  & 99.2 \tiny{.5} & 99.7 \tiny{.2} &
99.8 \tiny{.2}  & 99.6 \tiny{.3} & 99.5 \tiny{.3} & 99.2 \tiny{.4} & 99.0\tiny{.4} \\ %
\textbf{900} 
& 77.9 \tiny{2.5} & 
87.4 \tiny{2.2} &  
92.2 \tiny{1.7} & 
94.7 \tiny{1.8} & 
96.3 \tiny{1.2} &  
97.2 \tiny{.9} &
97.8 \tiny{.9}  &
98.5 \tiny{.5} & 
99.0 \tiny{.5} &
99.3 \tiny{.3}  & 
99.5 \tiny{.3} &
99.5 \tiny{.3} &
99.5 \tiny{.4} & 99.3\tiny{.5} & 99.2\tiny{.6} & 99.0\tiny{.8} & 
98.8 \tiny{.9}\\ %
\textbf{1000} & 
73.4 \tiny{2.2} & 
85.6 \tiny{1.5} &  
92.4 \tiny{1.5} & 
94.8 \tiny{1.2} & 
97.1 \tiny{.9} & 
98.7 \tiny{.5} & 
99.2 \tiny{.5} & 
99.5 \tiny{.4}  & 
99.6 \tiny{.5} & 
99.8 \tiny{.3} &
99.8 \tiny{.2}  & 
99.9 \tiny{.1} & 
99.1 \tiny{.1} & 
99.1 \tiny{.1} & 100.0\tiny{.1} & 
100.0\tiny{.1} & 
99.1 \tiny{.1} & 
100.0 \tiny{.1} & 100.0\tiny{.1} \\ \hline
& \textbf{100} & \textbf{150} & \textbf{200} & \textbf{250} & \textbf{300} & \textbf{350} & \textbf{400} & \textbf{450} & \textbf{500} & \textbf{550} & \textbf{600} & \textbf{650} & \textbf{700} & \textbf{750} & \textbf{800} & \textbf{850} & \textbf{900} & \textbf{950} & \textbf{1000} \\ \hline
\end{tabular}
\begin{tablenotes}
      \item Note: Trained sample lengths $\ell$ (first column) and accumulated sample length at each intermediate output (last row) are in thousands. Variance specified next to mean \footnotesize{in smaller font}.
\end{tablenotes}
\end{threeparttable}
\end{minipage}
\end{table*}

Tables~\ref{tab:sf_lstm} and~\ref{tab:sj_lstm} demonstrate crucial model insights that the \altname architecture yields in conjunction with our loss function for SF and SJ. For each $\ell$ that the model is trained on (row), we show the testing accuracy observed across intermediate outputs $O_{i}$ and correlate it with the aggregate number of samples that provide information for this classification. %
For instance, $\ell=100$ and $d=20$ ($m$ always equals $d/2$) yields 9 LSTM cells and their dense outputs, where, e.g., the $3$-rd output $O_{3}$ avails information from its own cell-specific input of $20$ samples as well as the previous two cells' combined insight from $60$ samples (with overlaps), corresponding to essentially the first $40$ samples from the $\ell$-sized input sub-drive.
Note that we use consistent model configurations for all $\ell$, i.e., $d=50$K and $s=25$K. 
First, we see in Tables~\ref{tab:sf_lstm} and~\ref{tab:sj_lstm} that for each $\ell$ (row), accuracy increases from left to right, substantiating our intuition that increasing entropy information is available with every LSTM cell for a given lane. %
More significantly, we see that the accuracy is consistent across columns for a given cell. For instance, for both SF and SJ, training a model with $\ell=400$K yields a final classification accuracy of $97\%$ which is also yielded by the cells corresponding to $400$K (net) sample-length of models trained with $\ell=500$K and above. This is a powerful result that essentially allow us to eliminate hours of compute and labor involved in characterizing the entropy of new roads and deciding what $\ell$ represents a suitable tradeoff between classification accuracy and timeliness. Even between SF and SJ in our dataset, we see that the former yields $90\%$ mean accuracy for $\ell=100$K while the latter over $95\%$; however, the value of $\ell$ also directly sets the minimum distance a vehicle must drive on a lane before being able to use \altname. %
Instead of consequently training \altname on each variation of $\ell$, however, we can, as a result of Tables~\ref{tab:sf_lstm} and~\ref{tab:sj_lstm}, train on large sub-drive lengths and inspect the intermediate outputs to characterize the accuracy achievable by smaller sub-drive lengths in this road. %
This visibility into the LSTM layer also allows us to train one model and infer for various sub-drive lengths if needed. This can be done by extracting weights from the trained model, from the first to the $i$-th sub-segment, to construct a model that can infer on samples of length corresponding to the $i$-th cell.

\section{Discussion}
\label{sec:disc}
We now discuss other practical considerations in deploying \altname and how to address them. Promising venues of further research emerge from some of these factors.

\textbf{Acquiring Training Data in Practice.}  We can utilize existing and/or imminent vehicular infrastructure for \altname's training data collection. First, \altname may leverage crowd-sourcing of pavement condition monitoring solutions. While \textit{Pavement Condition Index (PCI)}~\cite{sj_pci,sf_pci,sd_pci,dc_pci} indices were collected manually by visual survey traditionally, recent efforts investigate automated solutions leveraging sensors in cars. For example, Michigan Department of Transportation~\cite{mdot,mdot1} and Google~\cite{google_pothole,google_pothole_patent} look into crowd-sourcing pavement condition information gathered by vehicles. Voters is a start-up that spun-off from academic project to leverage instrumented vehicles to collect road characteristics~\cite{voters,voters_web}. Second, \altname may utilize crowd-sourced information from high-end vehicles that have accurate lane-level knowledge from camera-based solutions such as Street View Cars~\cite{googlestreet} or Tesla AutoPilot~\cite{teslaautopilot}. Third, \altname may also make use of crowd-sourcing data from self-driving cars employed by Google, Uber, Lyft, or others~\cite{googleself,uberself,lyft_self,mcfarland_2017,snavely_2017}. The self-driving cars are already actively on the road and are projected to be even more pervasive in a few years. For example, Uber and Lyft's self-driving cars are especially attractive as these companies publicly disclosed plans to convert their cars to self-driving vehicles in matter of years~\cite{lyft_roadmap}. For example, as per a 2017 estimate~\cite{uber_lyft_num}, there are at least 45,000 Uber and Lyft cars in San Francisco, which today contribute to over 170,000 trips on an average weekday in San Francisco~\cite{uberlyft2019}, and self-driving cab services are expanding to other cities as well~\cite{hawkins_2017}. When these cars are converted to self-driving vehicles, we expect the effort of collecting training data to reduce dramatically.

While we rely on these methods described above to acquire ground truth about lane-specific road surface characteristics, note that our data synthesis mechanism \textit{reduces the volume of ground truth needed.} As shown in Section~\ref{sec:eval}, we train \altname on thousands of drive capturing various vehicle and drive-specific variations with just 10-20 original drives per lane, and achieve $100\%$ lane classification accuracy. 

Further, we note that almost any ground truth collected traces collected by these vehicles equipped to correlate the lane they are on with accelerometer measures they collect can be used for training \altname. In practical scenarios, these vehicles may drive only over a small portion of a lane before changing roads or changing lanes. Since by design, \altname trains on sub-drives of different start/end points within the road, as long as $\ell$ samples are collected in given road, that portion of the drive may be used for training \altname, regardless of the number of lane switches containing in this sample. This could, in-fact, be a consideration in tuning the parameter $\ell$.

\textbf{Required Volume of Training Data.} We provide some back of the envelope estimates for the amount of training data that may be required for \altname to learn to classify between lanes of a $1$~km road. At the sampling rate of $111$Hz that the model has trained on, between $4000$-$16000$ samples are collected by a vehicle in $1$~km with driving speeds in the range $20$-$120$~km/h. Since each accelerometer measurement is a signed float requiring $4$~bytes of storage, this 1~km of data requires between $16$-$64$~KB of storage. Provided we need $5000-10000$ drives over the route to train over this route (irrespective of whether these are original or synthesized), $80-640$~MB of data is used. We also observe that the size of our trained LSTM models is between 7-10~MB depending on the length of training samples (spanning different distances over the road).

\textbf{Adapting to Changing Road Conditions.} 
Road surface signatures may change over time as roads exhibit increasing wear or get patched or resurfaced, and hence \altname must continually update its road surface model. %
We realize that analogous concerns arise in speech recognition where aging or health conditions cause voice changes in a speaker's acoustic model and adaptation to these is highly desirable. New acoustic conditions are often encountered that were not present originally at training time. %
This is handled by \textit{dynamic speaker adaptive} models, wherein adaptation data is only incrementally available. As long as camera-equipped vehicles that provide ground truth periodically collect and annotate road data, the model may be re-trained as these new measurements come in. Incremental and transfer learning provide promising areas of exploration to address this need. 
Several methods have been proposed in speech literature for such incremental model learning include the insertion of new layers in the model that are speaker-adaptive~\cite{gemello2007linear, neto1995speaker, li2010comparison, swietojanski2014learning, wang2018empirical}. %
While out of scope for this work, we postulate that similar techniques might be applied to time-series accelerometer data herein to adapt to changing road surfaces. %

\textbf{Model Generalization.} In this work, we've captured vehicle and drive-specific variations via the proposed data augmentation process and studied its effectiveness in improving \altname performance (see Section~\ref{sec:dataaug} and Figure~\ref{fig:effectOfAug}). As seen, the model accuracy against the test data directly increases as a function of each type of additional variation introduced in the training set, indicating increasing model generalizability. %
We also conduct preliminary experiments to test whether \altname generalizes across vehicles. This is especially challenging since only two vehicles were used to collect the data. However, even when trained on drives from only the Volkswagen (the original ones and synthesized ones), \altname yields $83$\% mean classification accuracy on the test dataset of the same car, and $78\%$ mean accuracy on drives of the Subaru. The performance of neural networks depends critically on the quantity of data they are trained with and how well the training data represents the underlying distribution being modeled. To achieve nearly $80$\% accuracy on the drives of a car that the network has never trained on before, while being training on drives from just a single vehicle of a different make/model, indicates the effectiveness of our data augmentation techniques and provides good reason to expect even easier generalization to vehicle- and drive-related factors as the variety of cars captured in the dataset increases.

\section{Conclusion}
\label{sec:conc}
In this work, we present \altname, a deep LSTM neural network for accurate lane classification. \altname is based on the observation that even between adjacent lanes, road surfaces exhibit differing characteristics caused by naturally occurring anomalies such as patches, bumps, and cracks. We capture this road surface information via low-cost accelerometers and use the measured vertical displacements of the vehicle from the ground for our experiments. We train \altname on data collected over 60~km of driving (and more synthesized with our data augmentation process). We demonstrate that \altname can distinguish between adjacent lanes with $100\%$ accuracy across different cities (San Francisco and San Jose) and road conditions. The entropy or information in road surface increases with distance, resulting in better classification for longer distances driven. We formulate a novel loss function that captures this and allows us to characterize the entropy in new roads and deploy \altname easily in new regions.  Our model yields $100\%$ lane classification accuracy with $200$~meters of driving data, achieving over $90\%$ with just $100$~m (correspondingly to roughly one minute of driving). %
We extensively test the model against frequent lane changes as well and achieve high performance.

\balance
\bibliographystyle{ACM-Reference-Format}
\bibliography{bib}

\end{document}